\pdfoutput=1
\documentclass[11pt]{article}

\usepackage{booktabs}  
\usepackage{tabularx}   
\newcolumntype{Y}{>{\centering\arraybackslash}c}
\usepackage{algorithm} 
\usepackage{algorithmicx} 
\usepackage{algpseudocode} 
\usepackage{amsmath} 
\usepackage{amssymb} 
\usepackage{xcolor}
\usepackage{pifont}
\newcommand{\greencheck}{{\color{green!60!black}\ding{51}}}
\newcommand{\redcross}{{\color{red!70!black}\ding{55}}}

\usepackage[preprint]{acl}
\usepackage{times}
\usepackage{latexsym}
\usepackage[section]{placeins}
\usepackage[T1]{fontenc}
\usepackage[utf8]{inputenc}
\usepackage{makecell}
\usepackage{microtype}
\usepackage{inconsolata}
\usepackage{graphicx}
\usepackage{booktabs}
\usepackage{arydshln}
\usepackage{listings}

\title{GUIDE: Towards Scalable Advising for Research Ideas}

\author{Yaowenqi Liu$^*$, Bingxu Meng$^*$, Rui Pan$^*$, Yuxing Liu, Jerry Huang,\\ \bf Jiaxuan You, Tong Zhang
  \\
  University of Illinois Urbana-Champaign
  \\
  \texttt{\{yl140, bingxum2, ruip4, yuxing6, jerry8, jiaxuan, tozhang\}@illinois.edu}
  \\
\\
}

\begin{document}
\maketitle
\def\thefootnote{*}\footnotetext{Equal Contribution.}
\begin{abstract}
The field of AI research is advancing at an unprecedented pace, enabling automated hypothesis generation and experimental design across diverse domains such as biology, mathematics, and artificial intelligence. Despite these advancements, there remains a significant gap in the availability of scalable advising systems capable of providing high-quality, well-reasoned feedback to refine proposed hypotheses and experimental designs. To address this challenge, we explore key factors that underlie the development of robust advising systems, including model size, context length, confidence estimation, and structured reasoning processes. Our findings reveal that a relatively small model, when equipped with a well-compressed literature database and a structured reasoning framework, can outperform powerful general-purpose language models such as Deepseek-R1 in terms of acceptance rates for self-ranked top-30\% submissions to ICLR 2025. Moreover, when limited to high-confidence predictions, our system achieves an acceptance rate exceeding 90\% on the ICLR 2025 test set, underscoring its potential to significantly enhance the quality and efficiency of hypothesis generation and experimental design. The code is released at \url{https://github.com/HowardLiu0830/GUIDE-Research-Idea-Evaluation}.
\end{abstract}

\section{Introduction}

Large Language Models (LLMs) have demonstrated remarkable progress in tasks ranging from text generation to code synthesis~\citep{achiam2023gpt4}. Recently, their application to \textit{academic research assistance}—especially in providing feedback on scientific writing and research ideas—has garnered increasing attention. Systems such as Google's \textit{Co-Scientist}~\citep{gottweis2025towards} exemplify a broader shift toward \textit{agentic LLMs} capable of collaborating with human researchers to improve scientific workflows, from hypothesis generation to peer review~\citep{jin2024agentreview, tan2024peer}. This emerging capability holds significant promise for accelerating scientific discovery and democratizing research access.

Among these agentic tasks, one particularly impactful yet underexplored area is the development of \textit{LLM-based advising agents}, models designed to provide detailed, constructive, and \textit{hallucination-free} suggestions for academic papers. The goal of these systems is to emulate human advisors by identifying strengths and weaknesses in submissions, suggesting actionable improvements, and assigning quantitative evaluations. However, existing LLMs often struggle with review fidelity: they may produce inflated ratings, fail to identify methodological flaws, or hallucinate evaluations not grounded in the text~\citep{ye2024we, yu2024automated, yan2024evaluating}. These limitations stem from a lack of fine-grained supervision, domain-specific alignment, and proper advising-style training data.

\begin{table*}[t]
\centering
\resizebox{\textwidth}{!}{%
\begin{tabular}{lcccc}
\toprule
\textbf{System} & \textbf{Retrieval-Augmented} & \textbf{Modular Summarization} & \textbf{Rubric-Guided Alignment} \\
\midrule
MetaGen~\citep{bhatia2020metagen} & \redcross & \redcross & \greencheck \\
MReD~\citep{shen2021mred} & \redcross & \redcross & \redcross \\
ReviewRobot~\citep{wang-etal-2020-reviewrobot} & \greencheck & \redcross & \redcross \\
Reviewer2~\citep{gao2024reviewer2} & \redcross & \redcross & \redcross \\
CycleResearcher~\citep{weng2024cycleresearcher} & \redcross & \redcross & \greencheck \\ \midrule
\textbf{GUIDE} & \greencheck & \greencheck & \greencheck \\
\bottomrule
\end{tabular}%
}
\caption{Comparison of LLM-based peer review systems. While some systems (e.g., CycleResearcher) are a part of broader end-to-end scientific agents, this comparison focuses specifically on their review capabilities.}
\label{tab:related-works}
\end{table*}

To address these challenges, we propose a novel and scalable framework for generating reliable, constructive, and expert-aligned suggestions. Our system is built upon a compressed knowledge base of paper summaries and metadata, distilled from full-text scientific papers in the field of machine learning, which enables efficient and accurate retrieval through a retrieval-augmented generation (RAG) pipeline. Before hypothesis verification, the system retrieves dozens of relevant papers to provide rich external context. Furthermore, to ensure that our models produce high-quality feedback, we introduce a \textit{rubric-guided alignment} strategy that instructs LLMs to follow and apply evaluation criteria akin to those used in major natural language processing (NLP) conferences. 

However, even with clear rubrics and guidelines, LLMs still exhibit the tendency to produce overly favorable and superficial revision suggestions. To address this issue, Reward rAnked FineTuning (RAFT;~\citealp{dong2023raft}) is used to align an open-source LLM with expert review criteria and domain-specific literature. This alignment enables our model to generate detailed, rubric-grounded feedback, with a particular emphasis on methodological rigor and experimental soundness—areas often neglected by existing systems. The combination of aforementioned techniques gives rise to our advising system: \textbf{G}uidelines (Rubrics),
\textbf{U}nderstanding (Summarized),
\textbf{I}nformation Retrieval (RAG), \textbf{D}irection (Advising Improvement with RLHF),
and \textbf{E}xplanation (LLM reasoning), or \textbf{GUIDE} in short. As shown in Table~\ref{tab:related-works}, compared to previous systems, GUIDE is the only one to incorporate all three components of retrieval augmentation, modular summarization, and rubric-guided alignment, highlighting its unique design and broader capability compared to prior approaches.

To evaluate GUIDE, we conduct a controlled experiment using the ICLR 2016–2024 paper submissions dataset, demonstrating the systematic improvements of our methods by predicting the acceptances of ICLR 2025 submissions. Specifically, we adopt the metrics of Top-5\% Precision, Top-30\% Precision, and Recall, which respectively evaluate the system's ability to identify high-quality papers, acceptable papers, and retrieve good papers. Additionally, a customized reward model based on rank classification is employed to accelerate the system's intermediate alignment.

Empirical results show that our system fine-tuned on the Qwen-2.5-7B-Instruct backbone model, outperforms large general-purpose language models in terms of rating alignment with actual acceptance. Moreover, rubric-guided prompting focusing on novelty and significance reduces hallucinated content and leads to more grounded, constructive feedback.

Our contributions are summarized as follows:
\begin{itemize}
    \item \textbf{End-to-end hypothesis advisor}: We introduce an LLM-based system \textbf{GUIDE} that provides actionable suggestions for both \textit{research ideas} and \textit{experimental design}. Our advising system, GUIDE-7B, outperforms large general-purpose LLMs such as GPT-4o-mini~\citep{achiam2023gpt4} and DeepSeek-R1~\citep{guo2025deepseek} in terms of Top-30\% precision—a metric that measures the acceptance rate of the top 30\% of papers, as rated based on the suggested strengths and weaknesses. 
    \item \textbf{Scalable advising with modular summarization}: We show that summarizing different sections of the literature separately effectively mitigates the limited context-length issue in idea advising scenarios, allowing more relevant content to be retrieved and compared for advising. In particular, the abstract and methodology sections are shown to be the most important for evaluating the quality of a paper.
    \item \textbf{Rubric-guided alignment}: We demonstrate that integrating rubric-based instruction significantly enhances the reviewer expertise of LLMs and improves evaluation usefulness.
\end{itemize}

\section{Related Works}
\paragraph{Hypothesis Discovery in Scientific Research.}  
Recent progress in large language models has enabled their integration into early-stage scientific workflows, particularly in hypothesis generation and ideation~\citep{zhou2022least, ruan2024liveideabench, singhal2025toward}. While these models have demonstrated promise in producing plausible hypotheses \cite{yao2023tree, tu2024towards, ruan2024liveideabench}, significantly less attention has been devoted to \textit{hypothesis verification}, the task of evaluating whether hypotheses are substantiated, methodologically sound, and experimentally grounded~\citep{yang2022language, qiu2023phenomenal}. Our work advances this understudied area by proposing a rubric-guided framework that evaluates scientific claims in a manner aligned with human peer reviewers.

\citet{jonesopenai, ifargan2025autonomous, swanson2024virtual, saab2024capabilities, taylor2022galactica} have proposed retrieval-augmented generation (RAG;~\citealp{lewis2020retrieval}) techniques to improve LLMs' access to external knowledge during hypothesis assessment. However, their methods do not adequately address the compression of retrieved content, leading to inefficiencies in multi-document settings. We introduce a prompt-learning-based compression approach that distills full texts into progressively shorter representations (e.g., summaries, abstracts, and titles) enabling more scalable and interpretable RAG pipelines.

\paragraph{End-to-End AI Scientist Agents.}  
Recent systems such as \textit{Co-Scientist} and \textit{CycleResearcher} aim to operationalize the full scientific lifecycle via autonomous agents~\citep{gottweis2025towards, weng2024cycleresearcher, lu2024ai, xu2024good} from idea generation to paper drafting and reviewing. While promising in scope, these systems treat review and critique as peripheral components \citep{skarlinski2024language}. While systems such as CycleResearcher~\citep{weng2024cycleresearcher} simulate the research-review loop through reinforcement learning, their reviews are not explicitly aligned with conference specific rubrics and lack retrieval grounding. Our system focuses on producing high-quality critique using rubric supervision and retrieval from similar papers, offering more actionable feedback for scientific writing. This specialization allows us to outperform generalist agents in review-centric evaluations and better support iterative paper improvement.

\paragraph{Summary.}  
Our work lies at the intersection of scientific hypothesis verification, automated peer review, and modular AI scientist systems. Departing from approaches that produce surface-level critiques or aim for full-lifecycle coverage, we present a focused, retrieval-augmented, rubric-aligned system that generates structured, high-fidelity scientific feedback.

\section{Method}
\subsection{Problem Definition}
Hypothesis evaluation is a crucial component of AI for Science. Rather than performing a full paper scan, our task focuses on assessing the core research hypothesis using four summarized sections: abstract, claimed contributions, method description, and experimental setup. This approach is especially useful in the early stages of paper writing, when only the outline of research ideas and experimental designs is available.

\subsection{Data Collection \& Generation}\label{sec:method:data_collect}
To prepare a database for literature comparison, we collect data from ICLR conferences spanning 2016 to 2024, from the publicly available OpenReview platform. For each submission, we obtained paper metadata (e.g., title, authors, abstract), full-text PDFs, official reviewer comments, and author rebuttals. Using a custom-built data cleaning pipeline, we processed these raw inputs into a structured database suitable for downstream use in both our RAG framework and in RAFT post-training~\citep{dong2023raft}. To convert full paper PDFs into markdown-formatted text, we utilized the open-source tool MinerU~\citep{wang2024mineru}, which enables reliable text extraction and structural segmentation.

An essential step in our preprocessing pipeline is content compression through summary generation. To this end, we used \texttt{gpt-4.1-nano}, a cost-effective yet high-performing model from OpenAI~\citep{achiam2023gpt4}, to generate structured section-wise summaries for each paper (e.g., Introduction, Related Work, Methodology, Experiments). This summarization reduced the input length of each paper by approximately $16\times$, allowing us to incorporate substantially more context within the RAG input window, mitigating the token limit bottleneck and improving retrieval efficiency in downstream tasks.

\begin{figure}[htbp]
  \centering
  \includegraphics[width=0.9\columnwidth]{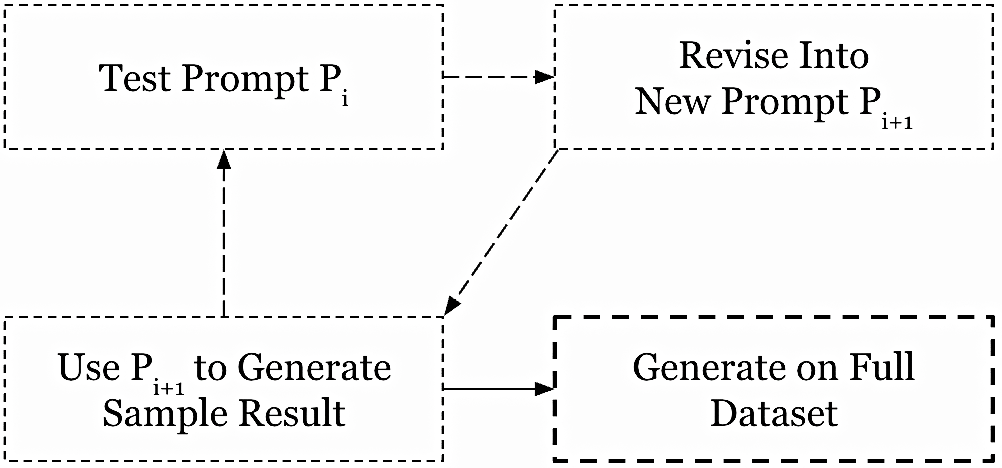}
  \caption{Contribution extraction with learned prompts.}
  \label{fig:prompt_eng_cycle}
\end{figure}

A crucial component of our data generation pipeline is \textit{contribution statement extraction} since contribution is considered the essence of a paper's strengths. This task is particularly challenging for two reasons: (1) not all papers explicitly state their contributions, and (2) such statements are rarely identifiable via simple rule-based or string-matching methods. To address this, we formalize the task as a sentence-level extraction problem over the full text of a paper in markdown format, where the objective is to identify explicit statements of the paper's key contributions. We assign a label of 1 if the contribution statement is explicitly present and correctly extracted, and 0 if the language model must infer it due to its implicit or absent formulation.

We employ OpenAI's cost-efficient model \texttt{gpt-4o-mini} to perform this extraction, leveraging self-consistency decoding and prompt optimization strategies as outlined in \citet{pan2024plumpromptlearningusing} and highlighted in Figure~\ref{fig:prompt_eng_cycle}. To systematically optimize our prompts, we construct a validation set of 100 human-annotated papers containing gold-standard contribution statements. We then apply a genetic algorithm to evolve prompts over successive generations. In each iteration, candidate prompts are scored based on the similarity between the extracted and ground-truth contribution statements, measured via metrics such as longest common subsequence (LCS) or Levenshtein distance. The top-$k$ prompts are selected and probabilistically propagated using Boltzmann-weighted sampling with $k_B T = 1$, guiding the evolutionary process toward higher-performing prompts across more than 100 trials. The resultant prompt achieves 94\% accuracy with an 80\% match based on Levenshtein distance or a 30\% match based on LCS. Examples of prompt evolution at both the initial and final iterations are provided in Appendix \ref{appendix:prompt_evolution}. A more detailed description of the genetic algorithm used for prompt engineering is also included.

Once the learned prompt is obtained, we employ it across all ICLR papers in our dataset. If a contribution statement is explicitly presented in the paper, it is labeled as 1; otherwise, it is labeled as 0. This automated step significantly improves scalability and accuracy over naive prompting or manual annotation, enabling high-quality contribution extraction at scale.

\subsection{Retrieval-Augmented Hypothesis Evaluator}
\begin{figure}[htbp]
  \centering
  \includegraphics[width=1.05 \columnwidth]{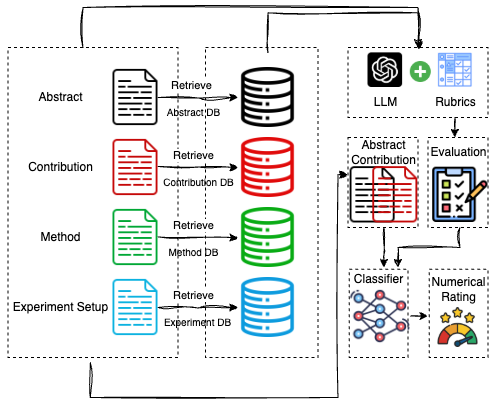}
  \caption{GUIDE: a RAG-based Advising System. The GUIDE pipeline begins by receiving the target paper's abstract, contribution, method, and experiment setup. For each of these sections, it retrieves corresponding content from a database of prior works. These target–exemplar pairs are fed into an LLM, which applies predefined rubrics to generate a structured advice. Finally, the idea's abstract and contribution together with this structured advice are passed to a lightweight classifier that produces the final numerical rating.}
  \label{fig:RAG}
\end{figure}

Accurately evaluating a research hypothesis requires more than just reading its abstract, claimed contributions, method description and experimental setup. It demands an understanding of how that specific hypothesis fits into the broader scientific literature. To address this, we design a rubric-guided RAG system. An illustration of our RAG pipeline is provided in Figure~\ref{fig:RAG}.
\label{sec:RAG}

\begin{figure*}[htbp]
  \centering
  \includegraphics[width=\textwidth]{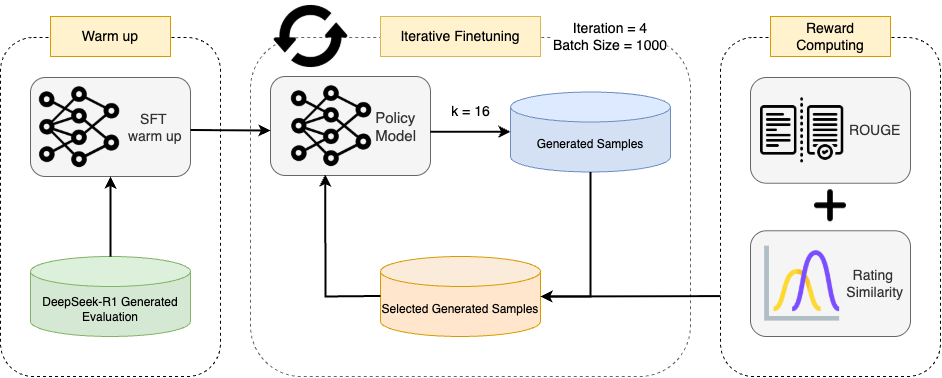}
  \caption{\textbf{Training Pipeline}: Overview of the two-stage training process for GUIDE-7B. \textbf{Stage 1 (Warming Up)} initializes GUIDE-7B with 4K idea-evaluation examples distilled from DeepSeek-R1, pairing each idea with rubric-based advice. \textbf{Stage 2 (RAFT)} further aligns GUIDE-7B with human evaluations by optimizing advice similarity (via ROUGE) and rating similarity (dot product of predicted and actual rating distributions). In RAFT, GUIDE-7B generates and selects the top-$k$ candidate advice responses for additional fine-tuning. Low-rated ideas encourage identification of weaknesses, while high-rated ideas prompt more positive feedback.}
  \label{fig:RAFT}
\end{figure*}

\paragraph{Retrieval Augmented System} We use OpenAI's \texttt{text-embedding-3-large} as our embedding model to build four separate databases, each storing abstract, claimed contributions, method description, and experimental setup respectively from 24,146 ICLR papers submitted between 2016-2024. During inference time, the system takes the target hypothesis's abstract, claimed contribution, method description, and experimental setup, and computes an embedding for each field. The system then queries each corresponding database to retrieve the top-$k$ most similar entries based on cosine similarity. The retrieved contexts are then appended to the original abstract, contribution, method, and experiment fields to form the full RAG context.

\paragraph{Rubrics Guided Prompting} During our experiments, we discovered that when evaluating hypothesis, existing LLMs tend to produce overly general feedback and fail to leverage the rich contextual information retrieved by our RAG pipeline. We address this by incorporating a set of evaluation rubrics into our system prompt that were extracted and distilled from the ICLR, ICML, and NeurIPS reviewing guidelines, with a focus on three core dimensions: novelty, significance, and soundness. We also instruct the LLM to partition its feedback into dedicated sections - one each for novelty, significance, and soundness. Each section contains focused commentary aligned with the corresponding rubric, and the final output conforms to our predefined JSON schema. For the full prompt and output specification, please refer to Appendix~\ref{appendix:prompt_format}

\subsection{Reward Ranked Fine-Tuning}
\label{sec:RAFT}
LLMs often suffer from implicit biases, leading to suboptimal or skewed outputs. In our experiments, we observed that off‐the‐shelf models tend to offer only positive and overly general praise and rarely provide neutral or critical feedback. This phenomenon highlights the need for better alignment with human evaluative standards. To improve the alignment of our models, we use Reward‐rAnked Fine‐Tuning (RAFT;~\citealp{dong2023raft}), an iterative fine-tuning algorithm with rejection sampling. Details of the pipeline are in illustrated in Figure~\ref{fig:RAFT}.

\paragraph{Warming Up} To empower general-purpose small LLMs to learn advising-centered reasoning and output formats, we adopt a warm-up phase at the start of training. Rubrics-prompted DeepSeek‐R1~\citep{guo2025deepseek} is employed to generate evaluations for a randomly sampled subset of ICLR 2024 papers, producing 4,000 high‐quality idea–evaluation pairs. These examples are used to perform an initial round of supervised fine‐tuning (SFT) of Qwen2.5‐7B‐Instruct.

\paragraph{Step 1: Generation}  
After warming up, RAFT~\citep{dong2023raft} is applied to further align the model with human preferences, which iteratively optimizes the model via generation, top-$k$ selection, and fine-tuning. At each iteration, the latest model generates \(K=16\) candidate advices for each of 1000 randomly selected ICLR 2024 hypothesis with experimental setups, where the top-$k$ candidate is selected ($k=1$) for each hypothesis.

\paragraph{Step 2: Best-of-$N$ Selection}  
For each candidate advice \(a_i\), we compute its rating distribution \(\hat{d}_i = [\hat{p}_{i,1}, \hat{p}_{i,2},\dots,\hat{p}_{i,10}]\in \mathbb{R}^{10}\) by concatenating the advice with the hypothesis's abstract and contributions and pass these contexts through our lightweight classifier, where \(\hat{p}_{i,j}\) denotes the probability of assigning rating \(j\) to the i-th hypothesis.   We construct the human reference distribution in two steps. First, given a set of observed human ratings \(\{r_k\}_{k=1}^K\) taking values in \(\{1,\dots,10\}\), the class counts are computed and and normalized into the distribution as follows:
\[
c_i = \bigl|\{k : r_k = i\}\bigr|,\quad
p_i = \frac{c_i}{\sum_{j=1}^{10}c_j},
\]
so that \(\sum_{i=1}^{10}p_i = 1\). Second, to avoid overly peaked distributions, we apply neighbor smoothing with coefficient \(\alpha\in[0,1]\), defining
\[
\tilde p_j =
\begin{cases}
(1-\alpha)p_1 + \alpha\,p_2, & j = 1,\\
(1-\alpha)p_j + \frac{\alpha}{2}\bigl(p_{j-1} + p_{j+1}\bigr), & 2 \le j \le 9,\\
(1-\alpha)p_{10} + \alpha\,p_9, & j = 10.
\end{cases}
\]

We denote the smoothed human distribution for hypothesis \(i\) as 
\[
\tilde d_i = [\tilde p_{i,1}, \tilde p_{i,2}, \dots, \tilde p_{i,10}].
\]
Here \(\tilde p_{i,j}\) is the smoothed probability of rating \(j\). Given the model's predicted distribution 
\(\hat d_i = [\hat p_{i,1}, \hat p_{i,2}, \dots, \hat p_{i,10}]\), 
the reward is calculated as
\[
R^{\mathrm{rating}}_i = \hat d_i \cdot \tilde d_i
    = \sum_{j=1}^{10} \hat p_{i,j}\,\tilde p_{i,j},
\]
where the form of weighted-sum avoids gradient vanishing issues in conventional softmax-based loss functions and encourages one-hot prediction.

To further reduce learning difficulty, we introduce an additional text‐similarity reward \(R^{\mathrm{text}}_i\) by measuring the ROUGE  score~\citep{lin2004rouge} between the generated advice and the concatenation of all reference human reviews.  The overall reward is then given by  
\[
R_i \;=\;\lambda\,R^{\mathrm{rating}}_i \;+\;(1-\lambda)\,R^{\mathrm{text}}_i,
\]
where \(\lambda\in[0,1]\) balances the two objectives.  We select the candidate advice with the highest \(R_i\) at each iteration for supervised fine‐tuning.  \\

\paragraph{Step 3: Fine-Tuning} After computing the combined reward, the advice $a_i^*$ with highest reward among $K$ candidates are selected, which forms a supervised fine‐tuning set \(\mathcal{S} = \{(x_i, a_i^*)\}_i\). Here \(x_i\) denote the retrieval augmented input for hypothesis $i$, i.e.\ the concatenation of the paper's abstract, claimed contributions, method description, experimental setup, and the retrieved summaries from Sec \ref{sec:RAG}. Qwen2.5‐7B‐Instruct is fine‐tuned on \(\mathcal{S}\).  By iterating this generate–select–fine‐tuning cycle, the model progressively learns to produce advices that maximize alignment with human judgments and textual fidelity.

\section{Experiment}
\subsection{Experiment Setting}
We build our retrieval databases using ICLR papers from 2016 to 2024, with a total of 24,146 valid papers. To prevent data leakage, we construct a test set of 1,000 papers randomly sampled from the set of ICLR 2025 submissions. Among these papers, 319 papers were accepted by the ICLR committee, which closely matches the conference acceptance rate of 31.7\%. To measure the advising system's alignment with human experts, the following metrics are adopted,

\begin{enumerate}
    \item \textbf{Top-5\% Precision:} Among all the hypotheses with the top-5\% highest predicted rating, the proportion that were actually accepted.
    \item \textbf{Top-30\% Precision:} Among all the hypotheses with the top-30\% highest predicted scores, the proportion that were actually accepted.
    \item \textbf{Accept Recall:} Among all the hypotheses that were accepted by ICLR 2025, the proportion that appear within the top 30\% predictions.

\end{enumerate}

\subsection{GUIDE-7B v.s. Deepseek-R1}
To validate the strong advising ability of GUIDE, we compare the predictiveness of its generated advice against that produced by general-purpose LLMs.
\paragraph{Setup}
\label{sec:exp1_setup}
We compare GUIDE with baselines using various large general-purpose LLMs, including GPT-4o-mini, QwQ-32B, and DeepSeek-R1, all equipped with retrieval-augmented generation, as described in Sec.~\ref{sec:RAG}. For all LLMs, the decoding hyperparameters are set to temperature $= 0.6$ and top-$p$ with $p=0.95$. Both GUIDE-7B and baselines will receive the input hypothesis's abstract, claimed contribution, method description, and experimental setup, along with the ten most relevant literature sections from our database.

\begin{table}[ht]
  \centering
  \small
  \resizebox{\linewidth}{!}{
  \begin{tabular}{lccc}
    \toprule
    \textbf{Model} &
    \shortstack{\textbf{Top-5\%}\\\textbf{Precision}} &
    \shortstack{\textbf{Top-30\%}\\\textbf{Precision}} &
    \shortstack{\textbf{Accept}\\\textbf{Recall}} \\
    \midrule
    GPT-4o-mini          & 70.0 $\pm$ 4.6\% & 47.7$\pm$ 2.4\% & 44.8$\pm$ 2.2\% \\
    QwQ-32B             & 66.7$\pm$ 1.2\% & 48.6$\pm$ 1.5\% & 45.8$\pm$ 1.4\% \\
    DeepSeek-R1          & 69.3$\pm$ 4.6\% & 50.2$\pm$ 0.5\% & 47.2$\pm$ 0.5\% \\
    \midrule
    GUIDE-7B         & \bf72.0$\pm$ 2.0\% & \textbf{51.3$\pm$ 0.4\%} & \textbf{48.3$\pm$ 0.3\%} \\
    \bottomrule
  \end{tabular}}
  \caption{Performance of advising systems with different backbones on the ICLR 2025 test set over three trials.}
  \label{tab:exp1}
\end{table}

\paragraph{Results}
As shown in Table~\ref{tab:exp1}, GUIDE-7B attains the highest Top-30\% Precision (51.3\%), outperforming all other variants. It is especially intriguing that GUIDE-7B, warmed up using datasets distilled from DeepSeek-R1, can surpass the original DeepSeek-R1. This improvement is largely attributed to the iterative RAFT alignment process, where GUIDE further learns from human preferences and acquires the ability to produce expert-level advice.

\subsection{Scalable Advising with Modular Summarization}
The compressed database also non-trivially contributes to the scalability of the system, as the retrieved content is summarized in shorter lengths to allow more literature to fit within the limited context window. To empirically verify this claim, we conduct ablation studies to compare the system's performance across different types of datasets.

\paragraph{Setup} For all comparisons, the input hypothesis is still formed by abstract, claimed contributions, method description, and experimental setup. The only difference lies in the different retrieval content.  All ablation runs use the same three backbone LLMs introduced in Section~\ref{sec:exp1_setup}: GPT-4o-mini, QwQ-32B, and DeepSeek-R1. We evaluate performance solely via the Top-30\% Precision metric on the held-out ICLR 2025 test set.  

\begin{table}[ht]
  \centering
  \small
  \resizebox{\linewidth}{!}{
  \begin{tabular}{lccc}
    \toprule
    \bf Retrieved Content            & \textbf{GPT-4o-mini} & \textbf{QwQ-32B} & \textbf{DeepSeek-R1} \\
    \midrule
    Full paper                            & 45.0\%               & 44.3\%           & 46.7\%               \\
    Abstract only                            & 47.0\%               & 45.7\%           & 49.0\%               \\
    + Contribution               & 46.7\%               & 46.0\%           & 48.7\%               \\
    + Method                              & \textbf{47.7\% }              & \textbf{48.7\%}           & 49.7\%               \\
    + Experiment                          & \textbf{47.7\%}               & 48.3\%           & \textbf{50.3\%}               \\
    \bottomrule
  \end{tabular}}
  \caption{Ablation on retrieved contents: Top-30\% Precision (\%) across different backbone LLMs.}
  \label{tab:exp2}
\end{table}

\paragraph{Results}
As shown in Table~\ref{tab:exp2}, summarization improves performance by allowing more relevant literature to be retrieved. The abstract and methodology turn out to be the two most conducive sources of information for advising, as the abstract naturally presents the main contribution of the paper, and the methodology contains objective information related to novelty and significance. One surprising observation is that experimental setups sometimes do not help. This is attributed to misaligned settings across the literature, where different works may use different setups to support their claims, while LLMs tend to prefer consistent setups.

\subsection{Rubrics Guided Prompting}
To quantify the effectiveness of rubric prompting and analyze the source of these potential improvements, we compare different rubrics under various backbones.

\paragraph{Setup}  
In this experiment, we fix the retrieved contents to be the same, all with 10 related abstracts, 10 contributions, 10 method summaries, and 10 experimental setups. The only difference across variants is the system prompt, which directs the LLM to emphasize a specific rubric (e.g., significance, novelty, soundness) and output the corresponding aspect‐focused evaluation. Since QwQ-32B exhibits relatively weaker instruction-following capabilities under prompting, it is  replaced with another widely adopted Gemini-flash-2.0 to ensure robust adherence to our system prompts.

\begin{table}[ht]
  \centering
    \small
  \resizebox{\linewidth}{!}{
  \begin{tabular}{lccc}
    \toprule
    \textbf{Prompts}    & \textbf{GPT-4o-mini} & \textbf{Gemini-flash-2.0} & \textbf{DeepSeek-R1} \\
    \midrule
    No rubrics         & 45.3\%               & 47.0\%           & 48.0\%               \\
    + Soundness only               & 44.7\%               & 43.3\%           & 47.3\%               \\ 
    + Novelty only                & 47.3\%               & 48.3\%              & 49.3\%               \\
    + Significance only            & 47.7\%                & 48.3\%           & 49.3\%               \\
  \midrule
    + All           & \textbf{47.7\%}               & \textbf{49.7\%}        & \textbf{50.3\%}               \\
    \bottomrule
  \end{tabular}}
  \caption{Top-30\% Precision (\%) with rubric prompts.}
  \label{tab:exp3}
\end{table}

\paragraph{Results}  
Significance and Novelty rubrics yield non-trivial gains in Top-30\% Precision, while Soundness guidance hurts performance. This phenomenon indicates that the general-purpose LLMs are still lacking the ability to assess experimental rigor. Overall, rubric prompts demonstrably enhance hypothesis evaluation, with the full system benefiting most from Significance and Novelty instructions.

\paragraph{Case Study}
As shown in Table~\ref{tab:exp3}, we observe that the significance rubric yields the most pronounced improvement in evaluation quality. To demonstrate the effectiveness of this rubric‐guided approach, a concrete example is presented to illustrate the model's assessment with and without significance rubrics applied to an ICLR2025 Oral hypothesis, as shown in Table~\ref{tab:case_study_rubrics}. For more detailed rubrics based prompting, please refer to Appendix~\ref{appendix:more-rubrics}.
\begin{table}[h]
  \small  
  \centering
  \noindent\textbf{Hypothesis:} \textit{Turning Up the Heat: Min-p Sampling for Creative and Coherent LLM Outputs}~\citep{nguyen2024turning} \textit{(ICLR2025 Oral)}
  \vspace{1ex}

  \begin{tabularx}{\columnwidth}{@{}%
      >{\raggedright\arraybackslash}p{0.23\columnwidth}%
      >{\raggedright\arraybackslash}X%
      Y@{}
    }
    \toprule
    & \textbf{Final Evaluation}  & \shortstack[c]{\textbf{Predicted}\\\textbf{Rating}} \\
    \midrule
    \textbf{With Significance Rubrics}
      & \dots It addresses a \textcolor{green!60!black}{well-known} problem in LLM decoding and offers a simple yet effective improvement over existing truncation methods, likely \textcolor{green!60!black}{to be adopted widely}.
      & 6.74 \\
    \cmidrule(lr){1-3}
    \textbf{No Rubrics Given}
      & \dots \textcolor{red!70!black}{While its empirical validation is thorough, the lack of theoretical grounding limits its conceptual novelty}.
      & 6.02 \\
    \bottomrule
  \end{tabularx}
  \caption{\textbf{Case Study:} Comparison of evaluation outcomes with and without significance rubrics for an ICLR2025 Oral hypothesis, demonstrating that rubrics guide the model to correctly identify high-impact contributions.}
  \label{tab:case_study_rubrics}
\end{table}

\subsection{Uncertainty Analysis}
Practical real-world applications normally demand high-confidence advising, which calls for further investigation into GUIDE's effectiveness under such conditions.
\paragraph{Setup} The model's uncertainty is quantified via the Shannon entropy of the predicted rating distribution:  
\[
H(\hat d) \;=\; -\sum_{j=1}^{10} \hat p_j \log \hat p_j.
\]  
A high entropy indicates that the classifier's probability mass is spread across many rating classes, whereas a low entropy reflects a focused, high‐confidence prediction.

To assess the impact of prediction confidence on evaluation accuracy, we rank all hypotheses by ascending entropy and select three subsets corresponding to the lowest‐entropy: 10\%, 20\%, and 30\% of papers.  Within each subset, we recompute the Top‐30\% Precision metric, measuring the proportion of truly accepted papers among the top 30\% of model‐ranked hypotheses.

\begin{figure}[htbp]
  \centering
  \includegraphics[width=1.05 \columnwidth]{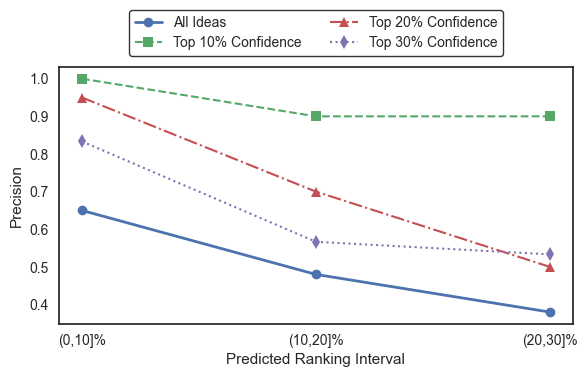}
  \caption{\textbf{Uncertainty Analysis}: Precision means the ratio of actually accepted papers over all papers that were within the specific confidence and predicted rating ranking interval. Predicted ranking interval means the set of papers sorted in descending order in terms of predicted rating.}
  \label{fig:uncertianty}
\end{figure}

\paragraph{Results} Fig.~\ref{fig:uncertianty}  reveals how model confidence modulates evaluation reliability: as the confidence and ranking threshold tightens, we generally observe higher precision, indicating that low‐entropy predictions are more trustworthy indicators of acceptance. Notably, within the top 10\% confidence subset, the Top-30\% Precision reaches the high accuracy of 93.3\%, suggesting that even for hypotheses not yet fully formalized into manuscripts, meeting both high‐confidence and high‐ranking criteria is \textit{a strong indicator of the final acceptance}. In automated hypothesis generation pipelines, where large batches of hypotheses can be generated at low cost, leveraging uncertainty enables the automatic selection of high‐quality hypotheses, underscoring the potential for accelerating research discovery.  

\section{Discussion}
Although our model and system are specifically tailored for academic hypotheses advising rather than full‐text academic paper review, in direct comparisons they outperform existing general-purpose LLM‐based baselines and achieve precision levels that closely approximate those of human reviewers.
\paragraph{Setup}
As a benchmark for human agreement, we consider the NeurIPS 2021 consistency experiment~\citep{beygelzimer2023has}, which assigns 10 percent of conference submissions to two independent review committees. It is important to note that the human baseline is drawn from the NeurIPS 2021 consistency experiment, which operated at a 24.5 \% acceptance rate—substantially lower than the 31.7 \% rate of ICLR 2025. As a result, GUIDE's performance should be even better, given that Top-24.5\% precision is strictly higher than Top-30\%. In addition, the human accuracy, precision, and recall reported here should be interpreted as a rough reference rather than a directly comparable benchmark.

For a strong and well‐known baseline, we employ the review agent component of AI Scientist~\citep{lu2024ai}. It processes the entire paper through multiple rounds of reflection and aggregates outputs via an ensemble of model passes to produce its final evaluation. By comparing against this full‐text, multi‐stage baseline, we demonstrate the advantages of our hypotheses‐centric advisor. Since GPT-4o-mini tends to reject all papers under AI Scientist's default settings, we replaced it with GPT-4.1-nano to ensure stable results.
\paragraph{Results}
\begin{table}[ht]
  \centering
  \small
  \resizebox{\linewidth}{!}{
  \begin{tabular}{lcc}
    \toprule
    \textbf{Baselines} & \textbf{Accuracy}& \textbf{F1} 
    \\  \midrule
    Human (NeurIPS)  & 73.4\%  & 48.4\% \\ 
     \hdashline\noalign{\vskip 0.5ex}
    AI Scientist with DeepSeek-R1  & 40.7\%   & 49.5\% \\
    AI Scientist with QwQ-32B & 42.7\% & 43.3\% \\
    AI Scientist with GPT-4.1-nano  & 61.2\%  & 20.8\%  \\
    GUIDE-7B  & \bf 69.1\%  & \bf 50.1\%\\ \bottomrule
  \end{tabular}}
  \caption{Performance comparison on the ICLR 2025 test set, where \textbf{Accuracy} (or Precision) $= \frac{\# \text{Correct Predicted Acceptances}}{\# \text{Predicted Acceptances}}$ represents the proportion of correct accept decisions over all accept decisions made by the system, and \textbf{F1} is the harmonic mean of Precision and Recall, with the Recall $= \frac{\# \text{Correct Predicted Acceptances}}{\# \text{Actual Acceptances}}$ being the proportion of correct accept decisions made by the system over all the actual accepted papers.}
  \label{tab:discussion}
\end{table}
Table \ref{tab:discussion} reveals that our idea‐centric advisor achieves similar performance as the human baseline in terms of acceptance rate. Moreover, our system outperforms the AI Scientist's review agent, which exhibits strong decision biases: GPT-4.1-nano accepts 17.1\% of papers, DeepSeek-R1 accepts 85.4\% of papers, and QwQ-32B accepts 69.2\%, all far away from the true 31.7\% rate. These results underscore the advantage of using a ranking-based evaluation standard, which more faithfully reflects selective thresholds and yields more balanced, reliable assessments.

\section{Conclusion}
Our study demonstrates that effective advising in hypothesis generation and experimental design does not necessarily require massive language models. By leveraging a compact model integrated with a compressed literature corpus and structured reasoning mechanisms, we achieve superior performance compared to larger, general-purpose models. The system's high acceptance rates, particularly on high-confidence predictions, highlight its practical utility in supporting scientific inquiry. These results suggest a promising path forward for building scalable, domain-aware advising systems that can meaningfully augment human creativity and decision-making in research.

\section*{Limitations}

This system currently focuses exclusively on the machine learning literature, allowing for more targeted retrieval, reasoning, and evaluation within a well-defined domain. By concentrating on a specific field, we are able to better optimize the summarization, advising, and alignment processes. However, extending the system to cover a broader range of scientific disciplines—such as biology, physics, or social sciences—remains an important direction for future work. Such expansion would require addressing additional challenges related to domain-specific terminology, varied writing styles, and diverse evaluation criteria.

\bibliography{custom}

\begin{thebibliography}{38}
\providecommand{\natexlab}[1]{#1}

\bibitem[{Achiam et~al.(2023)Achiam, Adler, Agarwal, Ahmad, Akkaya, Aleman, Almeida, Altenschmidt, Altman, Anadkat et~al.}]{achiam2023gpt4}
Josh Achiam, Steven Adler, Sandhini Agarwal, Lama Ahmad, Ilge Akkaya, Florencia~Leoni Aleman, Diogo Almeida, Janko Altenschmidt, Sam Altman, Shyamal Anadkat, and 1 others. 2023.
\newblock Gpt-4 technical report.
\newblock \emph{arXiv preprint arXiv:2303.08774}.

\bibitem[{An et~al.(2024)An, Zhou, Zou, and Yang}]{an2024iot}
Tuo An, Yunjiao Zhou, Han Zou, and Jianfei Yang. 2024.
\newblock Iot-llm: Enhancing real-world iot task reasoning with large language models.
\newblock \emph{arXiv preprint arXiv:2410.02429}.

\bibitem[{Beygelzimer et~al.(2023)Beygelzimer, Dauphin, Liang, and Vaughan}]{beygelzimer2023has}
Alina Beygelzimer, Yann~N Dauphin, Percy Liang, and Jennifer~Wortman Vaughan. 2023.
\newblock Has the machine learning review process become more arbitrary as the field has grown? the neurips 2021 consistency experiment.
\newblock \emph{arXiv preprint arXiv:2306.03262}.

\bibitem[{Bhatia et~al.(2020)Bhatia, Pradhan, and Pal}]{bhatia2020metagen}
Chaitanya Bhatia, Tribikram Pradhan, and Sukomal Pal. 2020.
\newblock Metagen: An academic meta-review generation system.
\newblock In \emph{Proceedings of the 43rd International ACM SIGIR Conference on Research and Development in Information Retrieval}, pages 1653--1656.

\bibitem[{Dong et~al.(2023)Dong, Xiong, Goyal, Zhang, Chow, Pan, Diao, Zhang, Shum, and Zhang}]{dong2023raft}
Hanze Dong, Wei Xiong, Deepanshu Goyal, Yihan Zhang, Winnie Chow, Rui Pan, Shizhe Diao, Jipeng Zhang, Kashun Shum, and Tong Zhang. 2023.
\newblock Raft: Reward ranked finetuning for generative foundation model alignment.
\newblock \emph{arXiv preprint arXiv:2304.06767}.

\bibitem[{Gao et~al.(2024)Gao, Brantley, and Joachims}]{gao2024reviewer2}
Zhaolin Gao, Kiant{\'e} Brantley, and Thorsten Joachims. 2024.
\newblock Reviewer2: Optimizing review generation through prompt generation.
\newblock \emph{arXiv preprint arXiv:2402.10886}.

\bibitem[{Gottweis et~al.(2025)Gottweis, Weng, Daryin, Tu, Palepu, Sirkovic, Myaskovsky, Weissenberger, Rong, Tanno et~al.}]{gottweis2025towards}
Juraj Gottweis, Wei-Hung Weng, Alexander Daryin, Tao Tu, Anil Palepu, Petar Sirkovic, Artiom Myaskovsky, Felix Weissenberger, Keran Rong, Ryutaro Tanno, and 1 others. 2025.
\newblock Towards an ai co-scientist.
\newblock \emph{arXiv preprint arXiv:2502.18864}.

\bibitem[{Guo et~al.(2025)Guo, Yang, Zhang, Song, Zhang, Xu, Zhu, Ma, Wang, Bi et~al.}]{guo2025deepseek}
Daya Guo, Dejian Yang, Haowei Zhang, Junxiao Song, Ruoyu Zhang, Runxin Xu, Qihao Zhu, Shirong Ma, Peiyi Wang, Xiao Bi, and 1 others. 2025.
\newblock Deepseek-r1: Incentivizing reasoning capability in llms via reinforcement learning.
\newblock \emph{arXiv preprint arXiv:2501.12948}.

\bibitem[{Ifargan et~al.(2025)Ifargan, Hafner, Kern, Alcalay, and Kishony}]{ifargan2025autonomous}
Tal Ifargan, Lukas Hafner, Maor Kern, Ori Alcalay, and Roy Kishony. 2025.
\newblock Autonomous llm-driven research—from data to human-verifiable research papers.
\newblock \emph{NEJM AI}, 2(1):AIoa2400555.

\bibitem[{Jin et~al.(2024)Jin, Zhao, Wang, Chen, Zhu, Xiao, and Wang}]{jin2024agentreview}
Yiqiao Jin, Qinlin Zhao, Yiyang Wang, Hao Chen, Kaijie Zhu, Yijia Xiao, and Jindong Wang. 2024.
\newblock Agentreview: Exploring peer review dynamics with llm agents.
\newblock \emph{arXiv preprint arXiv:2406.12708}.

\bibitem[{Jones(2025)}]{jonesopenai}
Nicola Jones. 2025.
\newblock Openai's' deep research'tool: is it useful for scientists?
\newblock \emph{Nature}.

\bibitem[{Lewis et~al.(2020)Lewis, Perez, Piktus, Petroni, Karpukhin, Goyal, K{\"u}ttler, Lewis, Yih, Rockt{\"a}schel et~al.}]{lewis2020retrieval}
Patrick Lewis, Ethan Perez, Aleksandra Piktus, Fabio Petroni, Vladimir Karpukhin, Naman Goyal, Heinrich K{\"u}ttler, Mike Lewis, Wen-tau Yih, Tim Rockt{\"a}schel, and 1 others. 2020.
\newblock Retrieval-augmented generation for knowledge-intensive nlp tasks.
\newblock \emph{Advances in neural information processing systems}, 33:9459--9474.

\bibitem[{Lin(2004)}]{lin2004rouge}
Chin-Yew Lin. 2004.
\newblock Rouge: A package for automatic evaluation of summaries.
\newblock In \emph{Text summarization branches out}, pages 74--81.

\bibitem[{Liu et~al.(2024)Liu, Ye, Xing, and Zou}]{liu2024reducing}
Sheng Liu, Haotian Ye, Lei Xing, and James Zou. 2024.
\newblock Reducing hallucinations in vision-language models via latent space steering.
\newblock \emph{arXiv preprint arXiv:2410.15778}.

\bibitem[{Lu et~al.(2024)Lu, Lu, Lange, Foerster, Clune, and Ha}]{lu2024ai}
Chris Lu, Cong Lu, Robert~Tjarko Lange, Jakob Foerster, Jeff Clune, and David Ha. 2024.
\newblock The ai scientist: Towards fully automated open-ended scientific discovery.
\newblock \emph{arXiv preprint arXiv:2408.06292}.

\bibitem[{Nguyen et~al.(2024)Nguyen, Baker, Neo, Roush, Kirsch, and Shwartz-Ziv}]{nguyen2024turning}
Minh Nguyen, Andrew Baker, Clement Neo, Allen Roush, Andreas Kirsch, and Ravid Shwartz-Ziv. 2024.
\newblock Turning up the heat: Min-p sampling for creative and coherent llm outputs.
\newblock \emph{arXiv preprint arXiv:2407.01082}.

\bibitem[{Pan et~al.(2024)Pan, Xing, Diao, Sun, Liu, Shum, Pi, Zhang, and Zhang}]{pan2024plumpromptlearningusing}
Rui Pan, Shuo Xing, Shizhe Diao, Wenhe Sun, Xiang Liu, Kashun Shum, Renjie Pi, Jipeng Zhang, and Tong Zhang. 2024.
\newblock \href {https://arxiv.org/abs/2311.08364} {Plum: Prompt learning using metaheuristic}.
\newblock \emph{Preprint}, arXiv:2311.08364.

\bibitem[{Qiu et~al.(2023)Qiu, Jiang, Lu, Sclar, Pyatkin, Bhagavatula, Wang, Kim, Choi, Dziri et~al.}]{qiu2023phenomenal}
Linlu Qiu, Liwei Jiang, Ximing Lu, Melanie Sclar, Valentina Pyatkin, Chandra Bhagavatula, Bailin Wang, Yoon Kim, Yejin Choi, Nouha Dziri, and 1 others. 2023.
\newblock Phenomenal yet puzzling: Testing inductive reasoning capabilities of language models with hypothesis refinement.
\newblock \emph{arXiv preprint arXiv:2310.08559}.

\bibitem[{Ruan et~al.(2024)Ruan, Wang, Hong, Wang, Liu, and Sun}]{ruan2024liveideabench}
Kai Ruan, Xuan Wang, Jixiang Hong, Peng Wang, Yang Liu, and Hao Sun. 2024.
\newblock Liveideabench: Evaluating llms' scientific creativity and idea generation with minimal context.
\newblock \emph{arXiv preprint arXiv:2412.17596}.

\bibitem[{Saab et~al.(2024)Saab, Tu, Weng, Tanno, Stutz, Wulczyn, Zhang, Strother, Park, Vedadi et~al.}]{saab2024capabilities}
Khaled Saab, Tao Tu, Wei-Hung Weng, Ryutaro Tanno, David Stutz, Ellery Wulczyn, Fan Zhang, Tim Strother, Chunjong Park, Elahe Vedadi, and 1 others. 2024.
\newblock Capabilities of gemini models in medicine.
\newblock \emph{arXiv preprint arXiv:2404.18416}.

\bibitem[{Shen et~al.(2021)Shen, Cheng, Zhou, Bing, You, and Si}]{shen2021mred}
Chenhui Shen, Liying Cheng, Ran Zhou, Lidong Bing, Yang You, and Luo Si. 2021.
\newblock Mred: A meta-review dataset for structure-controllable text generation.
\newblock \emph{arXiv preprint arXiv:2110.07474}.

\bibitem[{Singhal et~al.(2025)Singhal, Tu, Gottweis, Sayres, Wulczyn, Amin, Hou, Clark, Pfohl, Cole-Lewis et~al.}]{singhal2025toward}
Karan Singhal, Tao Tu, Juraj Gottweis, Rory Sayres, Ellery Wulczyn, Mohamed Amin, Le~Hou, Kevin Clark, Stephen~R Pfohl, Heather Cole-Lewis, and 1 others. 2025.
\newblock Toward expert-level medical question answering with large language models.
\newblock \emph{Nature Medicine}, pages 1--8.

\bibitem[{Skarlinski et~al.(2024)Skarlinski, Cox, Laurent, Braza, Hinks, Hammerling, Ponnapati, Rodriques, and White}]{skarlinski2024language}
Michael~D Skarlinski, Sam Cox, Jon~M Laurent, James~D Braza, Michaela Hinks, Michael~J Hammerling, Manvitha Ponnapati, Samuel~G Rodriques, and Andrew~D White. 2024.
\newblock Language agents achieve superhuman synthesis of scientific knowledge.
\newblock \emph{arXiv preprint arXiv:2409.13740}.

\bibitem[{Swanson et~al.(2024)Swanson, Wu, Bulaong, Pak, and Zou}]{swanson2024virtual}
Kyle Swanson, Wesley Wu, Nash~L Bulaong, John~E Pak, and James Zou. 2024.
\newblock The virtual lab: Ai agents design new sars-cov-2 nanobodies with experimental validation.
\newblock \emph{bioRxiv}, pages 2024--11.

\bibitem[{Tan et~al.(2024)Tan, Lyu, Li, Gao, Wei, Ma, Liu, and Li}]{tan2024peer}
Cheng Tan, Dongxin Lyu, Siyuan Li, Zhangyang Gao, Jingxuan Wei, Siqi Ma, Zicheng Liu, and Stan~Z Li. 2024.
\newblock Peer review as a multi-turn and long-context dialogue with role-based interactions.
\newblock \emph{arXiv preprint arXiv:2406.05688}.

\bibitem[{Taylor et~al.(2022)Taylor, Kardas, Cucurull, Scialom, Hartshorn, Saravia, Poulton, Kerkez, and Stojnic}]{taylor2022galactica}
Ross Taylor, Marcin Kardas, Guillem Cucurull, Thomas Scialom, Anthony Hartshorn, Elvis Saravia, Andrew Poulton, Viktor Kerkez, and Robert Stojnic. 2022.
\newblock Galactica: A large language model for science.
\newblock \emph{arXiv preprint arXiv:2211.09085}.

\bibitem[{Tu et~al.(2024)Tu, Palepu, Schaekermann, Saab, Freyberg, Tanno, Wang, Li, Amin, Tomasev et~al.}]{tu2024towards}
Tao Tu, Anil Palepu, Mike Schaekermann, Khaled Saab, Jan Freyberg, Ryutaro Tanno, Amy Wang, Brenna Li, Mohamed Amin, Nenad Tomasev, and 1 others. 2024.
\newblock Towards conversational diagnostic ai.
\newblock \emph{arXiv preprint arXiv:2401.05654}.

\bibitem[{Wang et~al.(2024)Wang, Xu, Zhao, Ouyang, Wu, Zhao, Xu, Liu, Qu, Shang et~al.}]{wang2024mineru}
Bin Wang, Chao Xu, Xiaomeng Zhao, Linke Ouyang, Fan Wu, Zhiyuan Zhao, Rui Xu, Kaiwen Liu, Yuan Qu, Fukai Shang, and 1 others. 2024.
\newblock Mineru: An open-source solution for precise document content extraction.
\newblock \emph{arXiv preprint arXiv:2409.18839}.

\bibitem[{Wang et~al.(2020)Wang, Zeng, Huang, Knight, Ji, and Rajani}]{wang-etal-2020-reviewrobot}
Qingyun Wang, Qi~Zeng, Lifu Huang, Kevin Knight, Heng Ji, and Nazneen~Fatema Rajani. 2020.
\newblock \href {https://doi.org/10.18653/v1/2020.inlg-1.44} {{R}eview{R}obot: Explainable paper review generation based on knowledge synthesis}.
\newblock In \emph{Proceedings of the 13th International Conference on Natural Language Generation}, pages 384--397, Dublin, Ireland. Association for Computational Linguistics.

\bibitem[{Weng et~al.(2024)Weng, Zhu, Bao, Zhang, Wang, Zhang, and Yang}]{weng2024cycleresearcher}
Yixuan Weng, Minjun Zhu, Guangsheng Bao, Hongbo Zhang, Jindong Wang, Yue Zhang, and Linyi Yang. 2024.
\newblock Cycleresearcher: Improving automated research via automated review.
\newblock \emph{arXiv preprint arXiv:2411.00816}.

\bibitem[{Xu et~al.(2024)Xu, Xue, Sheng, Deng, Ding, Shen, Fu, Wang, and Zhou}]{xu2024good}
Yi~Xu, Bo~Xue, Shuqian Sheng, Cheng Deng, Jiaxin Ding, Zanwei Shen, Luoyi Fu, Xinbing Wang, and Chenghu Zhou. 2024.
\newblock Good idea or not, representation of llm could tell.
\newblock \emph{arXiv preprint arXiv:2409.13712}.

\bibitem[{Yan(2024)}]{yan2024evaluating}
Ziyou Yan. 2024.
\newblock Evaluating the effectiveness of llm-evaluators (aka llm-as-judge). eugeneyan. com.

\bibitem[{Yang et~al.(2022)Yang, Dong, Du, Cheng, Cambria, Liu, Gao, and Wei}]{yang2022language}
Zonglin Yang, Li~Dong, Xinya Du, Hao Cheng, Erik Cambria, Xiaodong Liu, Jianfeng Gao, and Furu Wei. 2022.
\newblock Language models as inductive reasoners.
\newblock \emph{arXiv preprint arXiv:2212.10923}.

\bibitem[{Yao et~al.(2023)Yao, Yu, Zhao, Shafran, Griffiths, Cao, and Narasimhan}]{yao2023tree}
Shunyu Yao, Dian Yu, Jeffrey Zhao, Izhak Shafran, Tom Griffiths, Yuan Cao, and Karthik Narasimhan. 2023.
\newblock Tree of thoughts: Deliberate problem solving with large language models.
\newblock \emph{Advances in neural information processing systems}, 36:11809--11822.

\bibitem[{Ye et~al.(2024)Ye, Pang, Chai, Chen, Yin, Xiang, Dong, Shao, and Chen}]{ye2024we}
Rui Ye, Xianghe Pang, Jingyi Chai, Jiaao Chen, Zhenfei Yin, Zhen Xiang, Xiaowen Dong, Jing Shao, and Siheng Chen. 2024.
\newblock Are we there yet? revealing the risks of utilizing large language models in scholarly peer review.
\newblock \emph{arXiv preprint arXiv:2412.01708}.

\bibitem[{Yu et~al.(2024)Yu, Ding, Tan, Luo, Weng, Gong, Zeng, Cui, Han, Sun et~al.}]{yu2024automated}
Jianxiang Yu, Zichen Ding, Jiaqi Tan, Kangyang Luo, Zhenmin Weng, Chenghua Gong, Long Zeng, Renjing Cui, Chengcheng Han, Qiushi Sun, and 1 others. 2024.
\newblock Automated peer reviewing in paper sea: Standardization, evaluation, and analysis.
\newblock \emph{arXiv preprint arXiv:2407.12857}.

\bibitem[{Zhong et~al.(2024)Zhong, Zhang, Wang, Hou, Xiong, Zhu, Chen, Tan, Bi, Lewis et~al.}]{zhong2024law}
Ming Zhong, Aston Zhang, Xuewei Wang, Rui Hou, Wenhan Xiong, Chenguang Zhu, Zhengxing Chen, Liang Tan, Chloe Bi, Mike Lewis, and 1 others. 2024.
\newblock Law of the weakest link: Cross capabilities of large language models.
\newblock \emph{arXiv preprint arXiv:2409.19951}.

\bibitem[{Zhou et~al.(2022)Zhou, Sch{\"a}rli, Hou, Wei, Scales, Wang, Schuurmans, Cui, Bousquet, Le et~al.}]{zhou2022least}
Denny Zhou, Nathanael Sch{\"a}rli, Le~Hou, Jason Wei, Nathan Scales, Xuezhi Wang, Dale Schuurmans, Claire Cui, Olivier Bousquet, Quoc Le, and 1 others. 2022.
\newblock Least-to-most prompting enables complex reasoning in large language models.
\newblock \emph{arXiv preprint arXiv:2205.10625}.

\end{thebibliography}
\onecolumn

\appendix
\newpage
\raggedbottom
\section{Prompts and Output Format}\label{appendix:prompt_format}

\subsection{Inference}
In this section, we provide the detailed prompt and also the structured output JSON format.
\begin{table}[H]
  \small  
  \centering
  \begin{tabularx}{\textwidth}{|p{2.5cm}|X|}
    \hline

    Prompt Details          & SYSTEM:\\
                            & \texttt{``You are a professional hypothesis evaluator with expertise in machine learning.Your task is to evaluate a given target academic hypothesis step by step, with a focus on novelty, contribution and soundness. You will be given:  
1. The hypothesis's title, abstract, claimed contribution, method description, and experimental setup.
2. A set of relevant prior works, each with abstract, claimed contribution, method descriptions and experimental setups.
**Review Guidelines**
Read the given idea's content: It's important to carefully read through the given content, and to look up any related work and citations that will help you comprehensively evaluate it. Be sure to give yourself sufficient time for this step.**Evaluation Criteria** 1. Motivation / Objective: What is the goal of the paper? Is it to better address a known application or problem, draw attention to a new application or problem, or to introduce and/or explain a new theoretical finding? A combination of these? Different objectives will require different considerations as to potential value and impact. Is the approach well motivated, including being well-placed in the literature?
2. Novelty \& Originality: Are the tasks or methods new? Is the work a novel combination of well-known techniques? (This can be valuable!) Is it clear how this work differs from previous contributions?
3. Significance \& Contribution: Are the questions being asked important? Does the submission address a difficult task in a better way than previous work? Would researchers or practitioners likely adopt or build on these ideas?
4. Soundness: Can the proposed method and experimental setup properly substantiate the claimed contributions? Will the claims be well supported under the proposed experimental setup? Are the methods used appropriate? Is this a complete piece of work or work in progress?**Related-Works Usage**
1. **Abstract \& Contribution**: frame the problem, scope, and high-level ``what'' and ``why.'' Used for evaluating significance and novelty.
2. **Method**: describe ``how'' (algorithms, architectures and theoretical derivations). Used for checking whether the proposed method is novel or internally consistent, well-justified, and mathematically rigorous.
3. **Experimental setup**: specify experiment design, datasets, baselines, metrics. Used to evaluate whether this work's experiment is appropriately designed and whether the experiment is comprehensive enough to support the claims. This content may also contain expected results.**Criticality**
Noting the idea will become a paper submitted to top conferences with acceptance rate of 30\%, you should be more critical. Feel free to give negative evaluations if the idea's quality is poor.
For empirical works, there is no need to contain theoretical analysis. For theoretical works, there is no need to contain experimental. Do not give negative evaluations for the two cases.**Output Format**  
Provide a structured evaluation **strictly in valid JSON format**. Include both an overall evaluation and constructive suggestions for improvement. Do not add explanations, extra text, or Markdown formatting.  
When mentioning a related work, please use the title of the related work.''} \\
 & USER: \\
 & ``\texttt{**Title**: \dots **Abstract**: \dots **Claimed Contribution**: \dots**Method Description**: \dots **Experimental Setup**: \dots }
 
 \texttt{Below are the abstracts of key related works, outlining each study's scope and main findings. Use this section to evaluate the hypothesis's novelty and contributions: \dots}
 
 \texttt{Below are the key contributions of selected prior works, highlighting their novel ideas and advancements. Use this section to benchmark the new hypothesis's originality and impact: \dots}
 
 \texttt{Below are the methods of key related works, summarizing their technical approaches and algorithms. Use this section to assess the hypothesis's technical novelty, contribution, and soundness: \dots } 
 
 \texttt{Below are the experimental setups of key prior works, detailing their protocols and evaluation metrics. Use this section to judge whether the proposed experiments are sufficiently sound to support the hypothesis's claims: \dots}''\\
    \hline
    Output Format       & \texttt{\{``summary'': ``\dots'', ``comparison with previous works'': ``\dots'', ``novelty'': ``\dots'', ``significance'': ``\dots'', ``soundness'': ``\dots'', ``strengths'': ``\dots'',``weaknesses'': ``\dots'', ``evaluation'': ``\dots'', ``suggestion'': ``\dots''\}} \\
    \hline
  \end{tabularx}
  \caption{Prompt details and JSON output format}
  \label{tab:case_study}
\end{table}

\newpage
\subsection{Data Generation}\label{appendix:prompt_evolution}
In this section, we provide examples of evolving prompts for contribution extraction, with the following notable trends observed along the iterations of evolution: 
\begin{itemize}
    \setlength{\itemsep}{0em}
    \item Rubrics and guidelines emerge during the prompt evolution, which provide more specific instructions for LLMs
    \item More specific and detailed instructions emerge during the optimization, such as where to look for certain statements
    \item Prompt becomes more concise during the evolution.
\end{itemize}

\begin{table}[H]
\small
\centering
\begin{tabularx}{\textwidth}{|p{3.2cm}|X|}
\hline
\textbf{Final Prompt @ Iteration 28} &
\begin{minipage}[t]{\linewidth}
\texttt{You are an expert at extracting contribution statements from academic papers. Follow these instructions carefully:}

\vspace{0.5em}

\texttt{Task: Extract the contribution statements from the introduction section of the provided academic paper. Do not extract from the abstract.}

\vspace{0.5em}

\texttt{Guidelines:} \\
\texttt{- Exclude Abstract: Only consider the introduction section for extraction.} \\
\texttt{- Marked Contributions: If contributions are explicitly marked (e.g., bullet points, numbered lists), copy them exactly.} \\
\texttt{- Include All Points: Ensure every numbered or bulleted point is included.} \\
\texttt{- Unmarked Contributions: If not clearly marked, extract full contribution paragraph(s) near the end of the introduction.} \\
\texttt{- Headers and Phrases: Include headers or introductory contribution phrases.} \\
\texttt{- Single Source: Extract from one continuous block only.} \\
\texttt{- No Inference: Copy text verbatim unless inference is required for 0-label.} \\
\texttt{- Conciseness: Keep instructions clear and direct.} \\
\texttt{- Output Format:} \\
\texttt{\{"contribution\_label": <0 or 1>, "contribution\_text": "<text>"\}}

\vspace{0.5em}

\texttt{If contributions are found, set "contribution\_label": 1. If not, set "contribution\_label": 0 and provide a concise (at most 3 sentence) inferred summary.}
\end{minipage}
\\
\hline
\textbf{Initial Prompt @ Iteration 0} &
\begin{minipage}[t]{\linewidth}
\texttt{Extract the contribution statements from the introduction section of the given academic paper (exclude the abstract).}

\vspace{0.5em}

\texttt{- Copy Only Continuous Text:} \\
\texttt{  - Start at the beginning of the relevant paragraph.} \\
\texttt{  - Include the entire paragraph and its final sentence.} \\
\texttt{  - Include any headers or introductory phrases.}

\vspace{0.5em}

\texttt{- Output:} \\
\texttt{  - If contributions are found, set "contribution\_label": 1 and include the extracted text.} \\
\texttt{  - If not, set "contribution\_label": 0 and provide a concise (at most 3 sentence) inferred summary.}

\vspace{0.5em}

\texttt{- Do Not:} \\
\texttt{  - Search beyond the introduction.} \\
\texttt{  - Paraphrase or reword.} \\
\texttt{  - Hallucinate information.}

\vspace{0.5em}

\texttt{Return the result in JSON:} \\
\texttt{\{"contribution\_label": <0 or 1>, "contribution\_text": "<text>"\}}
\end{minipage}
\\
\hline
\end{tabularx}
\caption{Prompt evolution across iterations with structured output format}
\end{table}


\onecolumn
\section{End-to-End Example}\label{appendix:case_study}
Idea to evaluate: \textit{Law of the weakest link: Cross capabilities of large language models}~\citep{zhong2024law} (ICLR2025 Poster)

\begin{table}[H]
  \centering
  \begin{tabularx}{\textwidth}{|p{2.5cm}|X|}
    \hline
    \textbf{Setup} & \textbf{Evaluation} \\
    \hline
    GUIDE-7B (with rubrics) & \textbf{Summary}: Overall, the paper introduces CrossEval, a benchmark to evaluate cross-capabilities of LLMs, and identifies the 'Law of the Weakest Link,' showing that cross-capability performance is constrained by the weakest individual capability. It systematically defines cross-capabilities via taxonomies and conducts extensive evaluations on 17 LLMs.
    
      \textbf{Comparison with previous works}: Compared to previous works like CMMLU and SKILL-MIX, which focus on individual or combinatorial skills, this work uniquely addresses cross-capabilities. Unlike Corex or AgentBench, which emphasize multi-model collaboration or agent behavior, this paper emphasizes static, task-complexity analysis. While How Abilities in Large Language Models are Affected by Supervised Fine-tuning Data Composition studies capability scaling, this work highlights the importance of addressing weaker individual capabilities.
      
      \textbf{Novelty}: The paper introduces the 'Law of the Weakest Link' and a systematic framework for defining cross-capabilities via taxonomies. While prior works like GPT-Fathom evaluate LLM capabilities, this work focuses on cross-capability interactions, a novel angle.
      
      \textbf{Significance}: The findings highlight the critical role of addressing weaker individual capabilities to improve cross-capability performance, influencing model development priorities. The benchmark provides a comprehensive evaluation framework for future research.
      
      \textbf{Soundness}: The methodology uses expert annotations, multiple model responses, and LLM-based evaluators to ensure reliability. The experiments cover 17 models and 14 capabilities, with correlation analyses supporting claims. However, reliance on human judgments and limited model diversity (no open-source models) may introduce biases.
      
      \textbf{Strengths}: The strengths of the paper are the comprehensive taxonomy, expert-validated benchmark, systematic evaluation of cross-capability performance, and clear demonstration of the 'Law of the Weakest Link.' The inclusion of both static and dynamic (principle-based prompting) analyses strengthens the claims.
      
      \textbf{Weaknesses}: The weaknesses of the paper are the lack of diversity in evaluated models (all commercial), potential bias in taxonomy validation, and insufficient comparison with real-world tasks. The cross-capability definitions are abstract, and the experiments do not test cross-capability transfer to real-world scenarios.
      
      \textbf{Evaluation}: In conclusion, the paper makes a valuable contribution by systematically studying cross-capabilities and empirically validating the 'Law of the Weakest Link.' While the methodology is rigorous, broader model diversity and real-world validation would strengthen its impact.
      
      \textbf{Suggestion}: To improve the paper, the authors could expand model diversity to include open-source models, validate the taxonomy with real-world task datasets, and compare with benchmarks like AgentBench to demonstrate practical relevance.
    \\
    \hline
  \end{tabularx}
  \caption{Evaluation and suggestion generated by GUIDE-7B.}
\end{table}

\begin{table}[ht]
  \centering
  \begin{tabularx}{\textwidth}{|p{2.5cm}|X|}
    \hline
    \textbf{Setup} & \textbf{Evaluation} \\
    \hline
    DeepSeek-R1  (with rubrics) & \textbf{Summary}: Overall, the paper proposes the concept of cross capabilities in LLMs, introduces the CrossEval benchmark to evaluate them, and identifies the 'Law of the Weakest Link' phenomenon where cross-capability performance is constrained by the weakest individual capability. The methodology includes taxonomy construction, human-annotated benchmarks, and experiments across 17 models.
    
    \textbf{Comparison with previous works}: Compared to previous works like SKILL-MIX and L-Eval, which focus on skill combinations and long-context evaluation, this paper uniquely defines cross capabilities through structured taxonomies and emphasizes the bottleneck effect of weak individual skills. While Corex explores multi-model collaboration for complex tasks, this work focuses on intrinsic capability interactions within single models. SCALE and CMMLU address multilingual or domain-specific evaluations but do not systematically study cross-capability dependencies.
    
      \textbf{Novelty}: The paper introduces a novel framework for cross capabilities, combining predefined individual skills into paired tasks, and provides empirical evidence for the 'Law of the Weakest Link.' While skill combination is explored in SKILL-MIX, the structured taxonomy and weakest-link analysis are new contributions.
      
      \textbf{Significance}: The work highlights a critical limitation in LLMs' real-world applicability, offering a benchmark and actionable insights for improving weak capabilities. The findings could influence model development and evaluation practices, particularly in multi-dimensional tasks.
      
      \textbf{Soundness}: The methodology is rigorous, with 1,400 expert-annotated prompts, 8,400 human ratings, and evaluations across 17 models. Experiments include ablation studies and system-prompting interventions to validate claims. However, cross capabilities are limited to pairwise combinations, and the selection of individual capabilities (e.g., Spanish) lacks justification compared to core skills like reasoning.
      
      \textbf{Strengths}: The strengths of the paper are the systematic taxonomy of cross capabilities, the human-annotated CrossEval benchmark, the identification of the weakest-link phenomenon, and comprehensive experiments validating the findings across diverse models and evaluators.
      
      \textbf{Weaknesses}: The weaknesses of the paper are the limited scope of cross capabilities (only pairwise combinations), the inclusion of Spanish as a standalone capability without clear rationale, and insufficient comparison to SKILL-MIX's k-skill combination approach. The experiments do not explore mitigation strategies beyond system prompting.
      
      \textbf{Evaluation}: In conclusion, the paper makes a valuable contribution by formalizing cross capabilities and identifying a critical performance bottleneck in LLMs. While the benchmark and findings are impactful, the novelty is partially limited by prior work on skill combinations.
      
      \textbf{Suggestion}: To improve the paper, the authors could expand cross capabilities to include >2 skills, justify capability selection (e.g., Spanish vs. multilingual reasoning), compare with SKILL-MIX's combinatorial approach, and explore methods to mitigate the weakest-link effect beyond prompting.
    \\
    \hline
  \end{tabularx}
  \caption{Evaluation and suggestion generated by DeepSeek-R1.}
\end{table}

\twocolumn
\newpage
\section{Additional Experimental Results}
\subsection{The Effect of Different Databases}
To further evaluate the robustness of our system, we conduct additional experiments using the arXiv paper database for retrieval. Specifically, we construct a retrieval corpus comprising all papers in the cs.LG (machine learning) category up to December 2024. For each target paper, retrieval is restricted to papers published prior to the target, determined by the earlier of its ICLR submission date or the earliest arXiv posting. To prevent retrieval-level contamination, we also apply a ROUGE-L similarity check to ensure that the target paper itself is never retrieved as supporting context.

The evaluation results are shown in Table~\ref{tab:arxiv_retrieval}. GUIDE-7B, without any additional fine-tuning on the arXiv database, continues to outperform large-scale general-purpose language models, demonstrating its robustness to distribution shifts in the retrieval corpus.

\begin{table}[ht]
  \centering
  \small
  \resizebox{\linewidth}{!}{
  \begin{tabular}{lccc}
    \toprule
    \textbf{Model} &
    \shortstack{\textbf{Top-5\%}\\\textbf{Precision}} &
    \shortstack{\textbf{Top-30\%}\\\textbf{Precision}} &
    \shortstack{\textbf{Accept}\\\textbf{Recall}} \\
    \midrule
    Qwen2.5-7B-Instruct    & 60.0\% & 43.0\% & 40.4\% \\
    GPT-4o-mini          & 70.0\% & 48.0\% & 45.1\% \\
    QwQ-32B             & 68.0\% & 49.0\% & 46.1\% \\
    DeepSeek-R1          & 70.0\% & 51.0\% & 48.0\% \\
    \midrule
    GUIDE-7B         & \textbf{74.0\%} & \textbf{52.3\%} & \textbf{49.2\%} \\
    \bottomrule
  \end{tabular}}
  \caption{Performance of different models when retrieving from the arXiv cs.LG category.}
  \label{tab:arxiv_retrieval}
\end{table}
It is worth noticing that GUIDE-7B provides nearly 
10\% improvement compared to its base Qwen2.5-7B-Instruct model in almost all metrics. This demonstrates the significance of our training paradigm.

\subsection{The Quality Loss of Summarization}
To quantify the impact of modular summarization on advising performance, we conduct ablation experiments in which the number of retrieved papers is limited to 1–2, in order to control the total context length to 15k tokens, which is the same as in previous experiments. Table~\ref{tab:summarization_ablation} presents the Top-30\% Precision for each model when using either full papers or summaries as retrieved context.
\begin{table}[ht]
  \centering
  \small
  \resizebox{\linewidth}{!}{
  \begin{tabular}{lccc}
    \toprule
    \textbf{Model} &
    \shortstack{GPT-4o-mini} &
    \shortstack{QwQ-32B} &
    \shortstack{DeepSeek-R1} \\
    \midrule
    Full Paper    & 45.0\% & 44.3\% & 46.7\% \\
    Summary          & 44.7\% & 44.0\% & 46.7\% \\
    \bottomrule
  \end{tabular}}
  \caption{Top-30\% Precision with full paper vs. summary as retrieved context. The context length for full papers is limited to 15K tokens, and each summary corresponds to its full paper version during retrieval.}
  \label{tab:summarization_ablation}
\end{table}

As shown in Table~\ref{tab:summarization_ablation}, the performance loss from applying summarization is negligible across all models. This suggests that modular summarization successfully preserves the key aspects of contribution and novelty, while significantly reducing context length.

\subsection{Rating Distribution}
We also analyze the distribution of predicted ratings to assess whether LLM-based systems exhibit optimism bias compared to human reviewers.

\begin{table}[ht]
  \centering
  \begin{tabular}{lcc}
    \toprule
    Evaluator         & Mean & Variance \\
    \midrule
    GPT-4o-mini    & 6.40 & 0.10 \\
    QwQ-32B        & 6.17 & 0.26 \\
    DeepSeek-R1    & 6.04 & 0.19 \\
    GUIDE-7B       & 5.81 & 0.22 \\
    \hdashline\noalign{\vskip 0.5ex}
    Human          & 5.13 & 1.46 \\
    \bottomrule
  \end{tabular}
  \caption{Mean and variance of ratings assigned by human reviewers and different LLMs. GUIDE-7B significantly mitigates the optimism bias compared to existing general-purpose LLMs.}
  \label{tab:rating_mean_var}
\end{table}

Table \ref{tab:rating_mean_var} shows that human reviewers give the lowest average score (5.13) but with the greatest spread in their judgments (variance = 1.46). All LLM-based evaluators assign higher mean ratings, yet their variances (0.10–0.26) are much lower, indicating they rate more consistently but also more leniently. Among them, GUIDE-7B comes closest to human behavior, with a mean of 5.81 and a variance of 0.22.

To offer deeper insights into the behavior of different systems, we also provide a more granular visualization of the full rating distributions. By plotting the proportion of scores falling into each interval for both human reviewers and the model evaluators, the distinct rating behaviors and biases become immediately apparent and easier to interpret.

\begin{figure*}[t]
  \centering
  \includegraphics[width=0.32\linewidth]{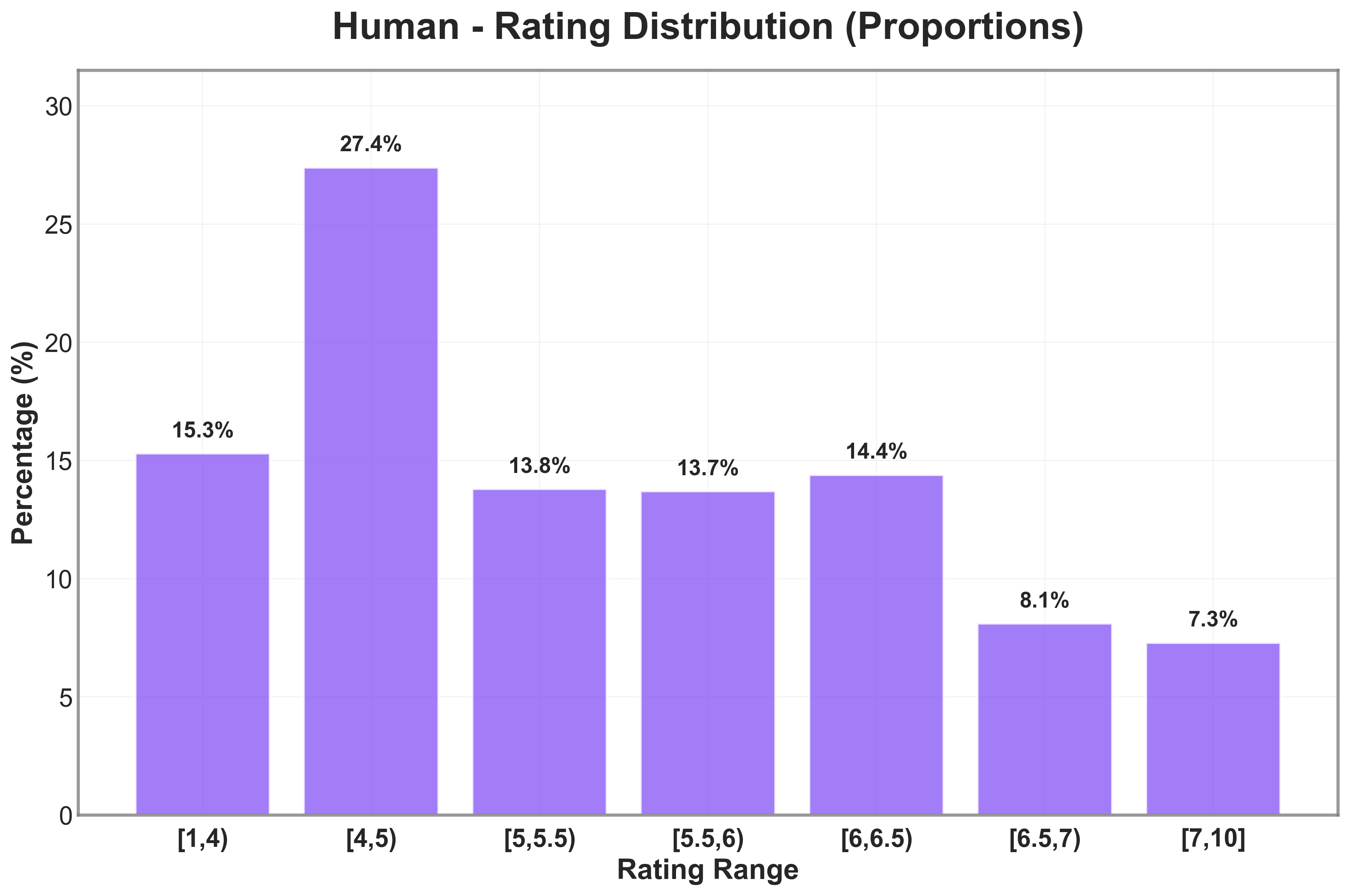}
  \includegraphics[width=0.32\linewidth]{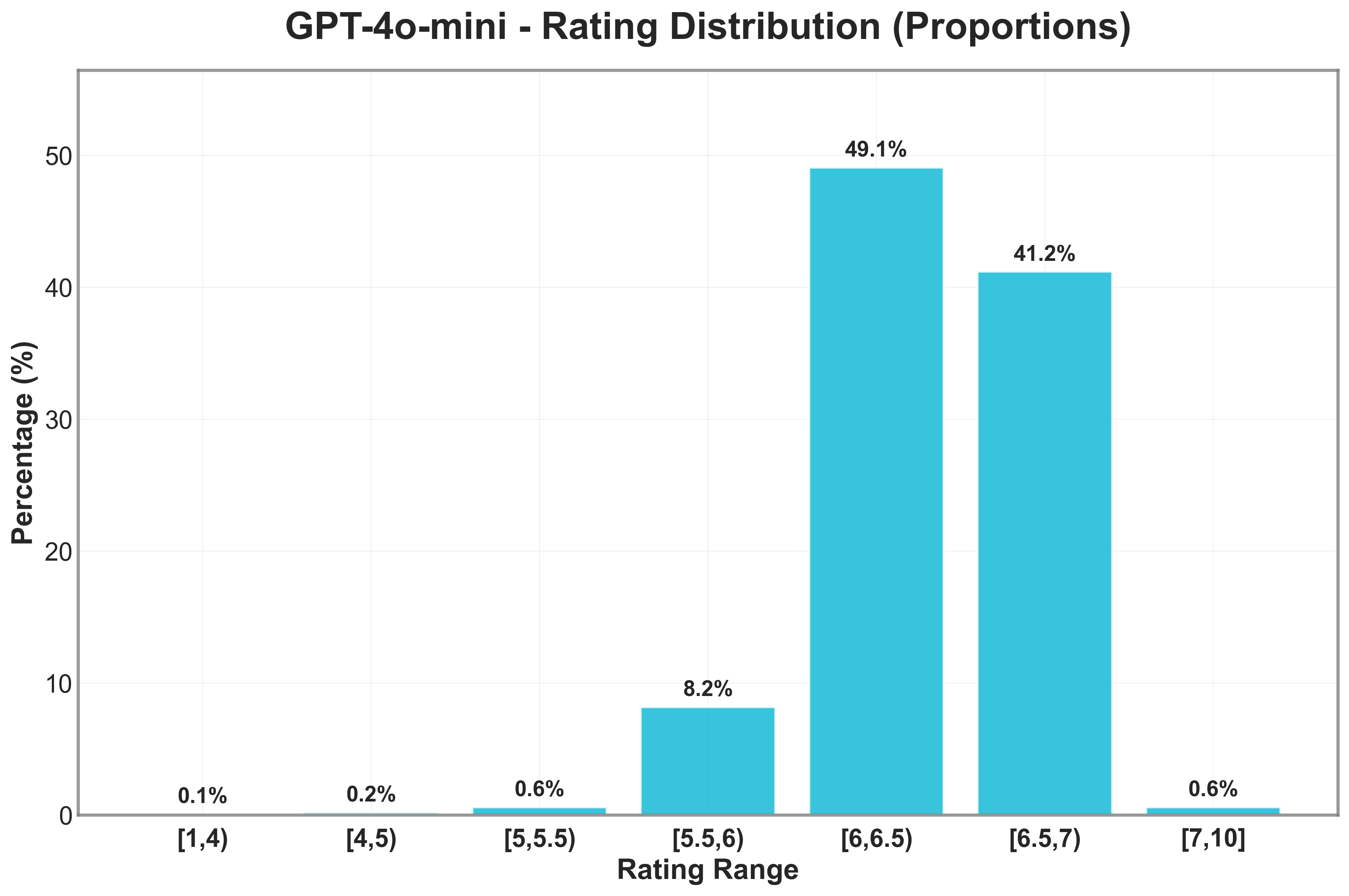}
  \includegraphics[width=0.32\linewidth]{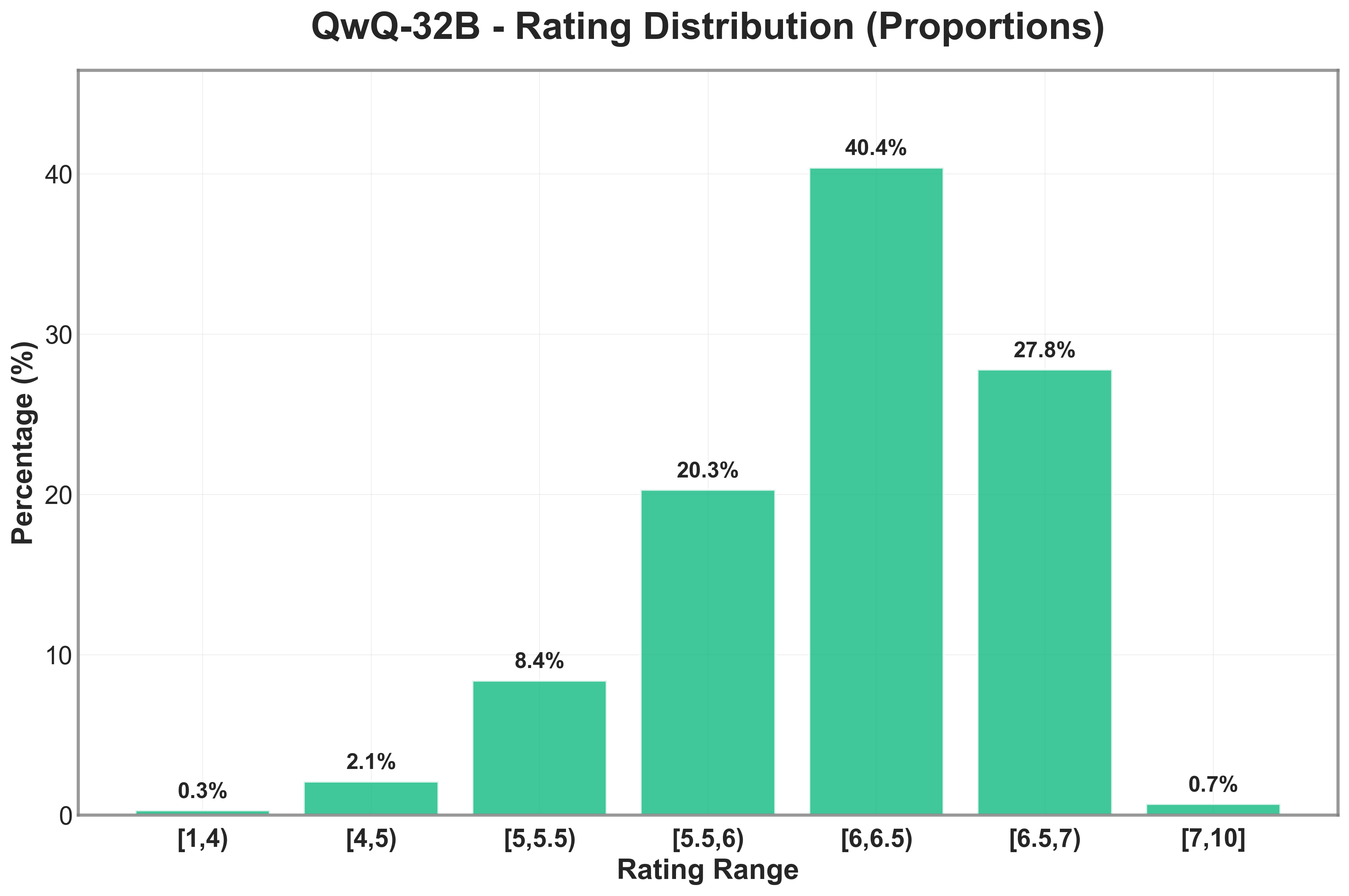}
  \includegraphics[width=0.32\linewidth]{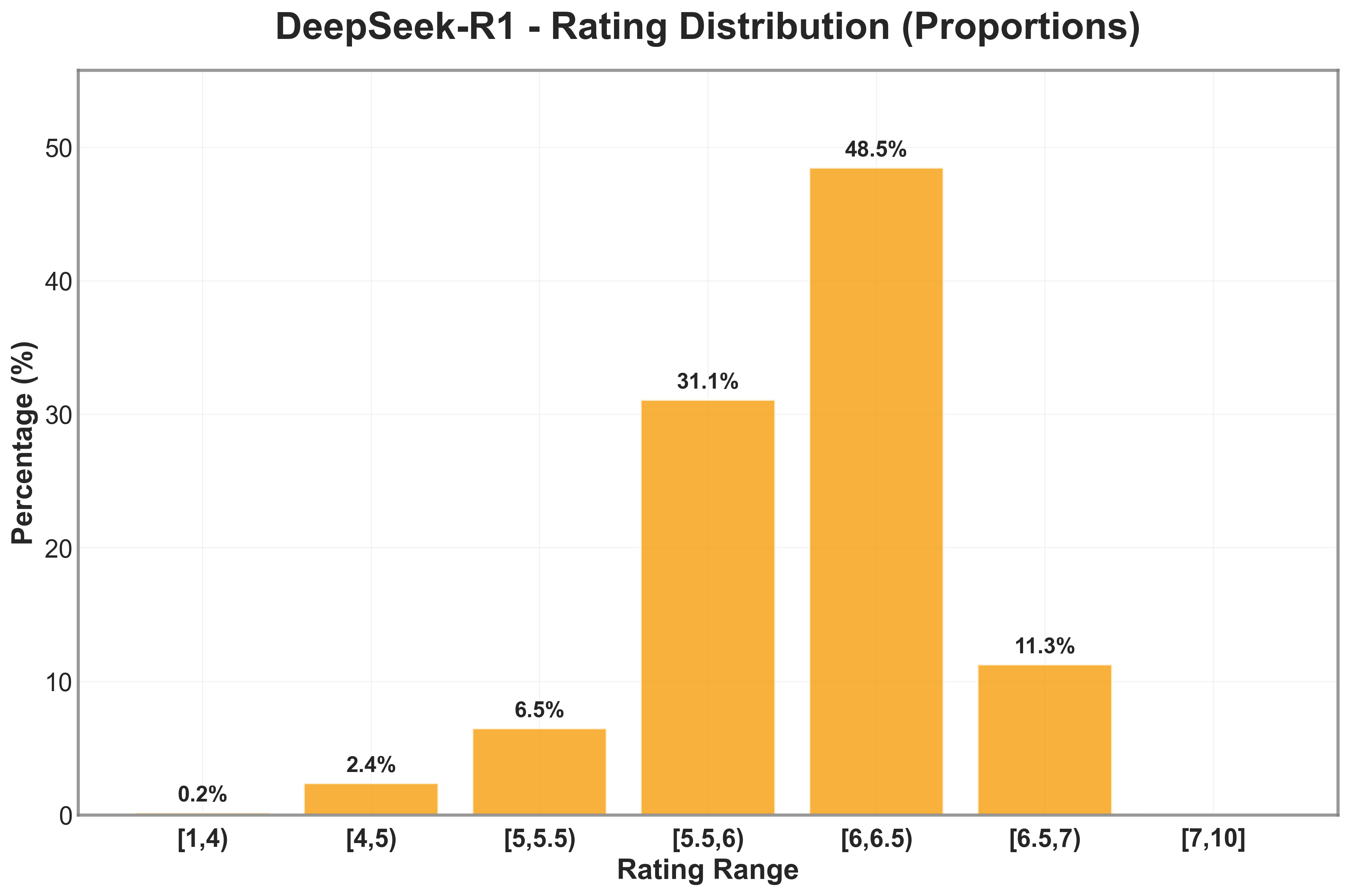}
  \includegraphics[width=0.32\linewidth]{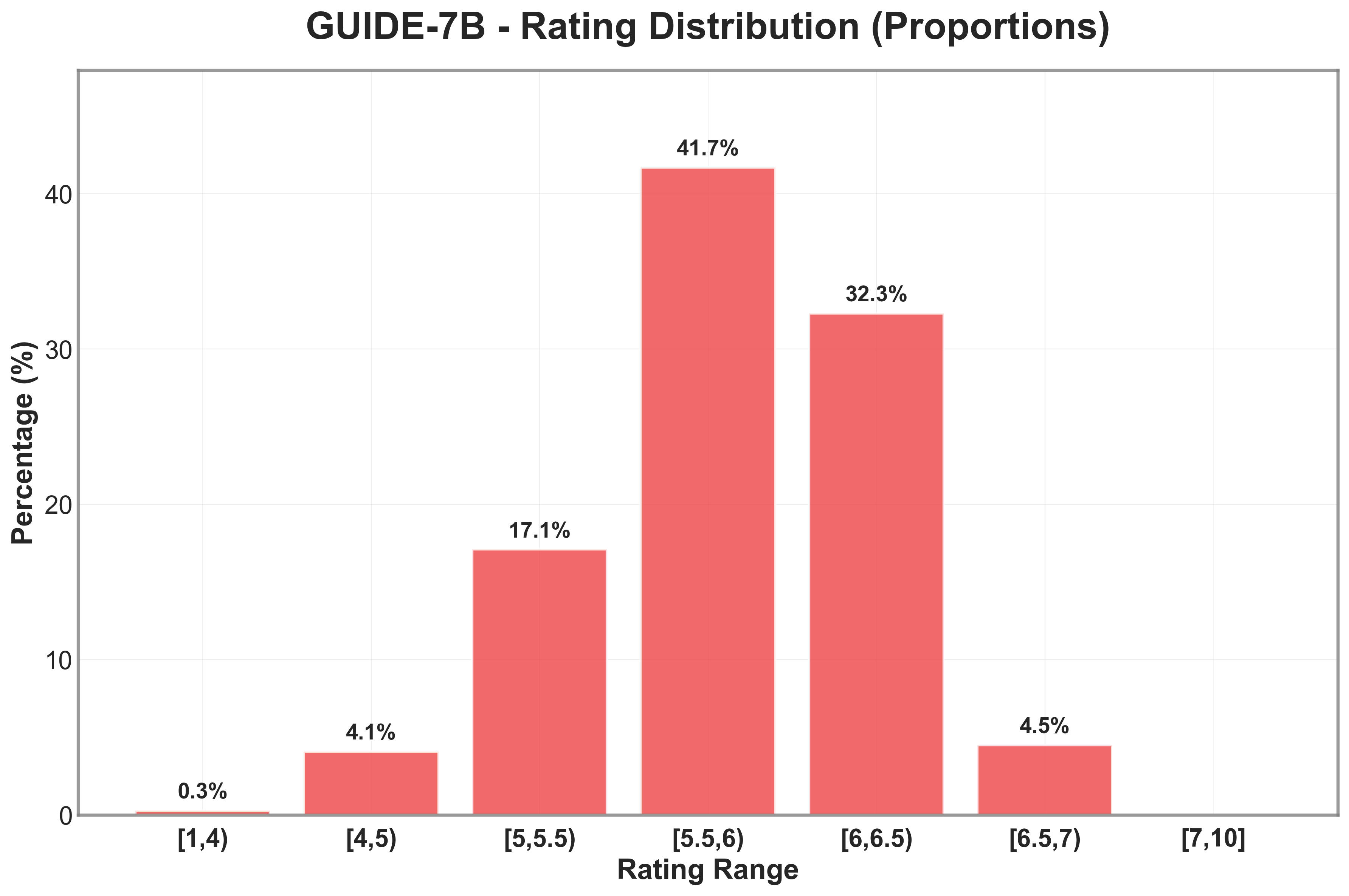}

  \caption{Distribution of ratings for human reviewers and various LLMs. General-purpose LLMs show optimism bias with ratings skewed toward higher intervals, while GUIDE-7B yields a more balanced, human-like rating distribution.}
  \label{fig:rating_distribution}
\end{figure*}

As illustrated in Table~\ref{fig:rating_distribution}, two key observations emerge. First, LLM-based systems exhibit a tendency to assign ratings that are concentrated near the mean, in contrast to human ratings, which are more evenly distributed across different intervals. Second, general-purpose LLMs such as GPT-4o-mini and QwQ-32B demonstrate a pronounced optimism bias, frequently assigning higher ratings in the [6.5, 7) range. In contrast, GUIDE-7B significantly mitigates this optimism bias, producing more balanced, borderline ratings in the [5.5, 6) interval for most papers. These findings highlight both the differences in rating behaviors between LLM-based systems and human evaluators, as well as the effectiveness of GUIDE-7B in reducing optimism bias.

\subsection{Rubrics Case Study}
\label{appendix:more-rubrics}
\begin{table}[h]
  \small  
  \centering
  \noindent\textbf{Title:} \textit{IoT-LLM: Enhancing Real-World IoT Task Reasoning with Large Language Models}~\citep{an2024iot}
  \vspace{1ex}

  \begin{tabularx}{\columnwidth}{@{}%
      >{\raggedright\arraybackslash}p{0.23\columnwidth}%
      >{\raggedright\arraybackslash}X%
      Y@{}
    }
    \toprule
    & \textbf{Final Evaluation}  & \shortstack[c]{\textbf{Predicted}\\\textbf{Rating}} \\
    \midrule
    \textbf{Human}
      & ``\dots I do not see substantial technical novelty in this study\dots the paper's novelty appears limited in this regard, \dots ''
      & 3.5 \\
    \midrule
    \textbf{Without Novelty Rubrics}
      & `` \dots The strengths of the paper are: 1) \textcolor{red!70!black}{Novel integration} of IoT data preprocessing with LLMs, addressing a critical gap in physical world reasoning\dots ''
      & 5.8 \\
    \midrule
    \textbf{With Novelty Rubrics}
      & `` \dots In conclusion, the paper presents a \textcolor{green!60!black}{moderately novel} application-focused framework with practical value in IoT-LLM integration \dots ''
      & 4.9 \\
    \bottomrule
  \end{tabularx}
  \caption{\textbf{Case Study:} Impact of novelty rubrics. Overall, novelty rubrics help prevent overrating papers with incremental contributions, thereby aiding in distinguishing strong papers from weak ones.}
  \label{tab:case_study_iot_llm}
\end{table}

\begin{table}[h]
  \small  
  \centering
  \noindent\textbf{Title:} \textit{Reducing Hallucinations in Large Vision-Language Models via Latent Space Steering}~\citep{liu2024reducing}
  \vspace{1ex}

  \begin{tabularx}{\columnwidth}{@{}%
      >{\raggedright\arraybackslash}p{0.23\columnwidth}%
      >{\raggedright\arraybackslash}X%
      Y@{}
    }
    \toprule
    & \textbf{Final Evaluation}  & \shortstack[c]{\textbf{Predicted}\\\textbf{Rating}} \\
    \midrule
    \textbf{Human}
      & ``\dots The analysis in the paper is thorough \dots''
      & 7.5 \\
    \midrule
    \textbf{Without Soundness Rubrics}
      & ``\dots In conclusion, the paper presents a \textcolor{green!60!black}{conceptually novel and empirically validated} approach to LVLM hallucination reduction \dots ''
      & 6.3 \\
    \midrule
    \textbf{With Soundness Rubrics}
      & ``\dots The idea is promising but \textcolor{red!70!black}{insufficiently validated}. While the approach is novel, the lack of comparison to state-of-the-art methods\dots''
      & 5.2 \\
    \bottomrule
  \end{tabularx}
  \caption{\textbf{Case Study:} Impact of soundness rubrics. Soundness rubrics can negatively impact the system's performance by imposing overly strict criteria on a paper's experimental design.}
  \label{tab:case_study_lvml_soundness}
\end{table}

\section{Dataset Analysis}
\begin{figure}[htbp]
  \centering
  \includegraphics[width=1.0 \columnwidth]{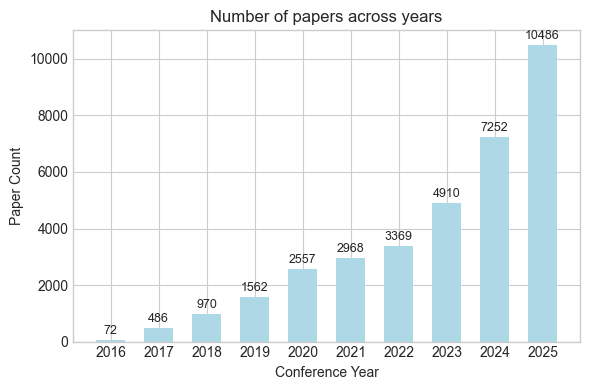}
  \caption{Annual ICLR paper counts from 2016 to 2025, illustrating rapid growth in recent years. Papers from 2016–2024 serve as our retrieval database.}
  \label{fig:paper_count}
\end{figure}

\paragraph{Dataset Size}
Our dataset comprises all available ICLR papers from 2016 through 2025 (34,632 papers in total). Of these, we adopt the 2016–2024 papers (24,146 papers) as our retrieval database. Figure~\ref{fig:paper_count} shows the annual publication counts.
\paragraph{Database Information}
We measured the token lengths of four sections: \emph{Abstract}, \emph{Claimed Contribution}, \emph{Method Description}, and \emph{Experimental Setup} across our retrieval database (24,146 papers) using the GPT-4 tokenizer (via the \texttt{tiktoken} library). Table~\ref{tab:section_lengths} reports the average token counts.

\begin{table}[ht]
  \centering
  \begin{tabular}{lc}
    \toprule
    Section                   & Avg. \# tokens \\
    \midrule
    Abstract                  & 229.0 \\
    Claimed Contribution      & 212.8 \\
    Method Description        & 201.9 \\
    Experimental Setup        & 193.0 \\
    \bottomrule
  \end{tabular}
  \caption{Average token lengths per section (GPT-4 tokenizer).}
  \label{tab:section_lengths}
\end{table}

\section{Experimental Details}

\subsection{Training}
\begin{figure}[htbp]
  \centering
  \includegraphics[width=\linewidth]{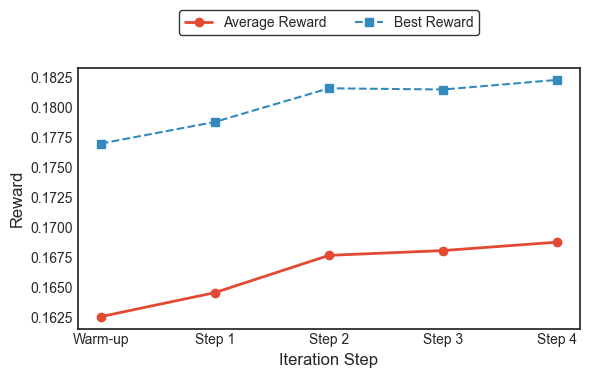}
  \caption{Average and best reward over successive RAFT iterations. Both metrics improve as training progresses, indicating the effectiveness of the RAFT pipeline for optimizing model performance.}
  \label{fig:reward_trend}
\end{figure}

We initialize GUIDE-7B from Qwen2.5-7B-Instruct and perform a two-stage training procedure. 
\paragraph{Warm-up:}For the warm-up stage, we use our RAG system with DeepSeek-R1 as the backbone to synthesize a high-quality dataset of $4,000$ samples. Training is carried out on 4x NVIDIA A100 40 GB GPUs with DeepSpeed's ZeRO 3 optimizations and CPU offload enabled. We train for 3 epochs using a batch size of 16, an initial learning rate of 1e-6 with a cosine decay schedule over a 15K context window.
\paragraph{RAFT:}
Our RAFT pipeline consists of three phases:

\begin{itemize}
  \item \textbf{Generation:} For each iteration, we use vLLM to sample 1,000 ICLR 2024 papers and generate \(K=16\) candidate evaluations per hypothesis with temperature \(0.7\), top-\(p\) \(=0.8\), and repetition penalty \(=1.05\).
  \item \textbf{Reward Computation:} We smooth human rating distributions with neighbor coefficient \(\alpha=0.4\) and compute the combined reward
    \[
      R_i = \lambda\,R^{\mathrm{rating}}_i + (1-\lambda)\,R^{\mathrm{text}}_i
      \quad\text{with}\quad \lambda=0.7.
    \]
    The text‐similarity reward \(R^{\mathrm{text}}\) is the sum of ROUGE‐1, ROUGE‐2, and ROUGE‐L, each weighted by 0.1 (total weight 0.3).
  \item \textbf{Supervised Fine-Tuning:} We keep fine-tuning our warmed-up model via LoRA with rank \(r=64\) and alpha \(=64\), using learning rate \(1\times10^{-5}\), batch size \(16\), for 2 epochs, and a cosine learning-rate schedule. We set the context window equal to 15k tokens. Training runs on 2 NVIDIA A100 40 GB GPUs with DeepSpeed's ZeRO 3 optimizations and CPU offload enabled.
\end{itemize}
\paragraph{Training Results:}
We ran 4 RAFT iterations. Figure~\ref{fig:reward_trend} shows how the average reward and best reward evolved over these iterations.

\subsection{Data Generation}

Details of the genetic algorithm's hyperparameter settings are:
\begin{itemize}
  \setlength{\itemsep}{0em}
  \item \textbf{Population:} All previous prompts form the population.
  \item \textbf{Fitness Function:} Accuracy based on 100 human-annotated gold labels (contribution detection correctness).
  \item \textbf{Selection:} Top-K (5) prompts are selected using weighted sampling.
  \item \textbf{Crossover (Recombination):} Performed by o1-mini to revise and combine prompts.
  \item \textbf{Mutation:} N/A — no random or absurd changes were introduced.
  \item \textbf{Iteration:} 28 iterations were performed before stopping.
\end{itemize}

All prompts are formulated as discrete, human-readable strings.

\section{Broader Impacts}
AI advisors can accelerate research progress by offering automated guidance, which supports researcher education and speeds up the development of new ideas. On the downside, the scoring capabilities of AI advisors could be misused in conference review processes. To prevent such misuse, we will release the scoring system only under appropriate regulatory frameworks.

\section{Human Annotation for Contribution Extraction}
The 100 annotated contributions in Sec.~\ref{sec:method:data_collect} were manually generated by a PhD student, who is one of the authors of this paper.

\section{AI Usage}
ChatGPT is used to correct grammatical errors and polish the paper writing.

\end{document}